\documentclass{article}

\usepackage{PRIMEarxiv}

\usepackage{float}
\usepackage[utf8]{inputenc} 

\usepackage[T1]{fontenc}    
\usepackage{url}            
\usepackage{booktabs}       
\usepackage{amsfonts}       
\usepackage{nicefrac}       
\usepackage{microtype}      
\usepackage{lipsum}
\usepackage{fancyhdr}       
\usepackage{graphicx}       
\usepackage{authblk}
\usepackage{subcaption}
\usepackage{hyperref}       
\graphicspath{{media/}}     
\usepackage[backend=biber,style=numeric,sorting=none,maxbibnames=3]{biblatex} 
\title{Vision-Language-Action Models: Experimental Insights from a Real-World UR5 Platform
}

\author {Mathilde Hochedel}
\author{Marc Lalonde}

\affil{  Luqia Technologies \\
  405 Ogilvy Ave., \#101 \\
  Montreal, QC, Canada\\
  \texttt{\{mathilde.hochedel, marc.lalonde\}@luqia.ca}\\
}

\begin{document}
\maketitle

\section*{Abstract}
This project investigates whether recent Vision-Language-Action (VLA) models can be transferred from controlled research benchmarks to a real-world robotic platform, specifically a UR5e manipulator, in a reproducible and operationally meaningful manner. The work integrates real-robot data acquisition, dataset engineering, data structuring that is compatible with the RLDS (Reinforcement Learning Datasets) format, and the fine-tuning and deployment of OpenVLA and OpenVLA-OFT models, alongside systematic validation of action representations and control interfaces. From an engineering perspective, the project resulted in several foundational assets: (i) a complete real-robot data acquisition pipeline, (ii) a dataset conversion workflow aligned with RLDS standards, (iii) an initial fine-tuning and inference infrastructure for VLA models, and (iv) a structured set of experimental observations grounded in real-robot trials. These elements collectively establish a reproducible framework for evaluating learning-based manipulation systems beyond simulation. Empirically, the experiments reveal a consistent gap between promising offline indicators and unstable closed-loop behavior on the physical system. Detailed analysis shows that this gap cannot be attributed solely to model limitations. Instead, it is strongly influenced by a combination of factors, including action semantics, coordinate frame conventions, temporal alignment between modalities, image preprocessing consistency, and dataset coverage and quality. These observations lead to a key interpretation: the successful deployment of VLA systems in real-world settings depends less on incremental improvements in model capacity and more on precise control of the entire data–model–control pipeline. In particular, small inconsistencies in data representation or system integration can propagate and significantly degrade performance at inference time. In conclusion, the project reframes VLA-based robotics from a primarily model-centric challenge to a system-level problem. The project highlights the difficulty of running robust task execution on the real robot and provides a clear, experimentally grounded understanding of the conditions required for reliable deployment. This work therefore establishes a practical foundation for future efforts, emphasizing the need for tighter integration between learning algorithms, control strategies, data acquisition protocols, and simulation-based iteration.

\section{Introduction}
\subsection{Project Motivation}
The project originates from a broader initiative in intelligent robotics aimed at assessing the practical relevance of recent Vision-Language-Action (VLA) models for manipulation tasks. More specifically, the goal was not to survey or benchmark all existing VLAs, but to conduct a focused experimental study on a small number of open and practically deployable models that could serve as a realistic entry point for real-robot investigation.

This motivation is based on two observations. First, VLA models offer a compelling paradigm: they directly map visual observations and natural-language instructions to robot actions, suggesting a strong potential for flexible task specification, rapid prototyping, and more natural human–robot interaction \cite{vla_survey, rt1, rt2}. Second, despite rapid progress in the field, the ecosystem remains fragmented, with limited methodological guidance for deploying and adapting open-source VLA models to new robotic setups under real-world conditions \cite{vla_survey}.

Within this landscape, the project deliberately focused on OpenVLA \cite{openvla} as its primary experimental baseline. This choice was not based on the assumption that OpenVLA is the best available VLA model, but rather on its specific relevance for research-oriented deployment. OpenVLA was explicitly introduced by its authors as an open-source generalist VLA intended to support fine-tuning and adaptation \cite{openvla, openvla_repo}, in response to the limited accessibility of prior VLA systems and the lack of practical recipes for adoption. As stated in the article, “robotics needs open-source, generalist VLAs that support effective fine-tuning and adaptation” (\cite{openvla}, Section Introduction), a position that aligns directly with the objectives of the present project. In addition, OpenVLA is distributed with open checkpoints, code, and a training pipeline, making it especially suitable for experimentation by external researchers \cite{openvla_repo}. A second motivation for this choice was the methodological continuity. Once OpenVLA had been selected as the initial baseline, OpenVLA-OFT \cite{openvla_oft} appeared as a natural extension: it revisits key OpenVLA design choices while remaining close enough in architecture and tooling to enable controlled progression in experimental complexity. This made it possible to investigate whether improvements in action representation and decoding strategy could be explored without rebuilding the entire pipeline around a fundamentally different model family.

Another observation that motivates the project is the significant gap between research settings and practical deployment conditions. Most existing work is validated in simulation or on specific academic platforms, with limited emphasis on transferability to collaborative industrial robots. In addition, VLA systems operate as reactive closed-loop policies: they iteratively predict and execute actions based on current observations. While this enables adaptability, it also creates strong sensitivity to the consistency of the entire perception–action loop, including sensing, preprocessing, control interfaces, and action semantics.

These observations lead to a central interpretation: evaluating VLA models in real-world conditions is not only a question of model quality but of system integration. The deployment of such models requires coordinated control over data acquisition, representation, software infrastructure, and hardware constraints. In this context, the project is motivated by the need to move from conceptual promise to operational understanding. More specifically, it seeks to answer a practical question: what is required, in terms of data, expertise, and system design, to reliably execute real-world manipulation tasks using open, fine-tunable VLA models?

\subsection{Research questions and hypothesis}
\label{sect:hyp}
Building on the motivation of the project, this work is structured around a set of research questions that target both model behavior and system-level constraints in real-world deployment. 

A first set of questions concerns the transferability of the model. Specifically, we investigate whether OpenVLA can be applied zero-shot from its original training domains to a local UR5e setup, and whether a relatively small fine-tuning dataset (consistent with recommendations from the original documentation \cite{openvla_repo}) is sufficient to adapt the model to this new environment. These questions reflect a larger uncertainty regarding the generalization potential of VLA models outside their reference contexts.

A second set of questions focuses on the role of data. Early observations suggest that performance may strongly depend not only on dataset size, but especially on acquisition protocol, demonstration variability, temporal structure, and preprocessing consistency. This leads to the question of whether data-related factors dominate model performance in real-world conditions.

These questions are supported by several working hypotheses derived from both the documentation and preliminary observations. First, it is hypothesized that the presence of idle steps in demonstrations may bias the model toward freezing or averaging behaviors, rather than producing goal-directed actions (\cite{openvla_repo}, Section VLA Performance Troubleshooting of the README file). Second, inconsistencies between image preprocessing during dataset construction and the internal preprocessing pipeline used during training and inference are expected to degrade performance. Third, ambiguities in action representation, particularly with respect to the reference frames and normalization/denormalization conventions, are hypothesized to significantly affect predicted actions, limiting generalization across setups. Finally, a methodological hypothesis is introduced: the consistency and quality of the dataset and pipeline can be partially validated using a simpler imitation learning baseline. If such a baseline exhibits similar failure modes, this would support the interpretation that limitations arise from data and system design rather than from the VLA model itself.

Taken together, these elements lead to a central interpretation guiding the study: real-world performance of VLA systems is likely to be governed less by high-level model capabilities than by precise alignment of data, representations, and control interfaces. The experiments are therefore designed to test this hypothesis through controlled comparisons and systematic analysis of failure modes.

\section{Background}

\subsection{Global Overview of Vision-Language-Action Models}
VLA models represent a recent paradigm shift in robotics, with the aim of unifying perception, reasoning, and control within a single end-to-end framework \cite{vla_survey, rt1}. Formally, a VLA model takes visual observations and natural language instructions as input and directly produces low-level robot actions, typically in the form of control commands \cite{rt1, rt2}. This contrasts with most common approaches (autonomous vehicles, for example) where perception, planning, and control were treated as separate modules.

From an architectural perspective, VLAs follow a structured yet integrated pipeline. As illustrated in the survey \cite{kawaharazuka2025vla_review}, the process can be decomposed into three main components: (1) a vision encoder that extracts features from images, (2) a language-aware backbone, typically based on large language models, that fuses visual and textual information and performs high-level reasoning, and (3) an action decoder that maps the resulting representation to continuous or discretized control signals. The complete pipeline observation followed by interpretation and action is therefore learned jointly, rather than explicitly engineered. This end-to-end formulation changes significantly from classical modular robotics pipelines, where perception (e.g., object detection), planning (e.g., task decomposition) and control (e.g., trajectory generation) are independently designed and optimized. Although modular systems offer interpretability and controllability, they are typically limited to predefined tasks and require extensive engineering to adapt to new scenarios. In contrast, VLAs aim to learn generalist policies capable of handling diverse tasks and instructions, leveraging large-scale pre-training and multimodal data \cite{rt2, octo}. The promise of this paradigm lies in its potential for scalability and generalization. By jointly modeling vision, language, and action, VLAs are expected to transfer knowledge across tasks and robotic platforms, reducing the need for task-specific data collection and enabling more flexible deployment. However, this promise remains partially fulfilled \cite{vla_survey,kawaharazuka2025vla_review}. Indeed, current observations highlight several open challenges. First, robustness and reliability in real-world environments remain limited, with most systems performing well only under constrained conditions \cite{kawaharazuka2025vla_review}. Second, generalization across domains and embodiments is still an open problem, particularly due to discrepancies between training data and deployment settings \cite{kawaharazuka2025vla_review}. Third, the end-to-end nature of VLAs introduces issues of interpretability, as the internal decision process is less transparent than in modular systems. Finally, the  feasibility of deployment is constrained by data requirements, computational cost, and the lack of standardized evaluation protocols.

These observations lead to a key interpretation: while VLAs offer a compelling framework for unifying perception and control, their practical effectiveness is currently limited by system-level constraints, particularly related to data, evaluation, and real-world variability. Consequently, assessing their performance requires not only model-level analysis, but also a careful examination of the full experimental pipeline, from data acquisition to control execution.

\subsection{OpenVLA}
\begin{figure}[H]
    \centering    
    \includegraphics[width=12cm]{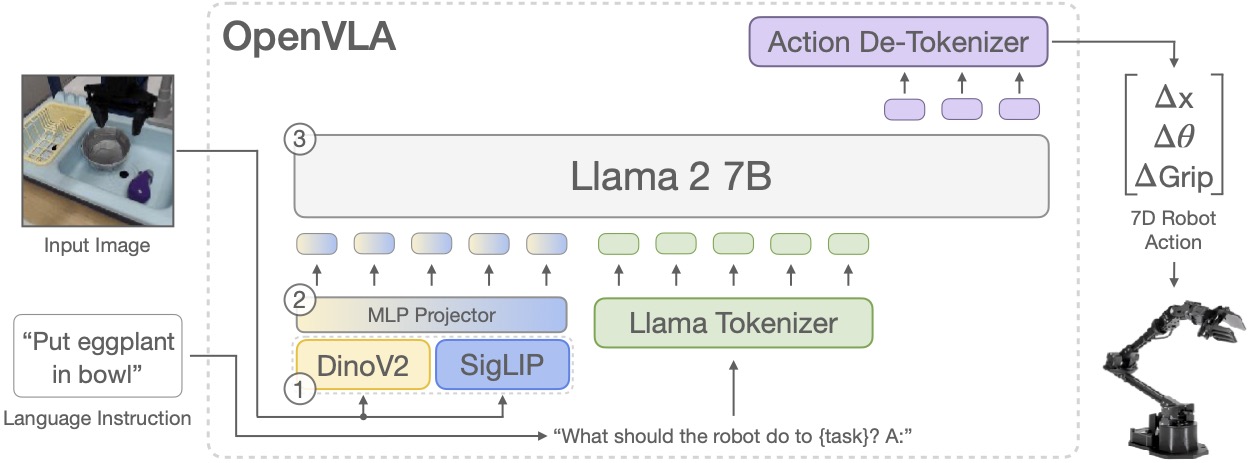}
    \caption{Architecture of the OpenVLA model, as proposed in \cite{openvla}}
    \label{fig:openvla-arch}
\end{figure}

OpenVLA relies on a standard Vision-Language Model (VLM) architecture composed of three main components (1) a visual encoder, (2) a projection module, and (3) a Large Language Model (LLM) backbone as shown in Figure~\ref{fig:openvla-arch}. More specifically, the visual encoder combines two pretrained models, DINOv2 and SigLIP, whose features are concatenated. This dual-encoder design is motivated by the need for robust spatial understanding in robotic manipulation. In particular, DINOv2 contributes improved spatial reasoning capabilities, which are critical in third-person camera setups where depth and perspective ambiguities arise. The visual features are then projected into the embedding space of the LLM via a lightweight MLP, enabling multimodal fusion. The backbone itself is based on LLaMA 2 (7B), which processes both the encoded visual tokens and the tokenized language instruction to generate output. This design reflects a key principle of VLA models: rather than designing task-specific perception and control modules, OpenVLA leverages a general-purpose multimodal backbone. Spatial reasoning, language understanding, and action prediction are implicitly learned within a unified representation space, benefiting from large-scale pretraining. OpenVLA can thus be viewed as a direct extension of VLMs to robotics, where the core architectural innovation lies not in new modules, but in re-purposing existing multimodal models for control.

A central challenge in OpenVLA is that the LLM backbone is inherently designed to generate text tokens, whereas robot control requires continuous-valued actions. To bridge this gap, OpenVLA discretizes each dimension of the action space into 256 bins, producing integer-valued tokens for each action dimension. These discrete tokens are then mapped into the LLaMA tokenizer vocabulary by overwriting the 256 least-used tokens, effectively repurposing them to represent robot actions. The model is trained using a standard next-token prediction objective, where only the action tokens contribute to the loss. At inference time, the predicted tokens are de-tokenized and mapped back to continuous control commands, typically followed by a de-normalization step to match the physical scale of the robot. This formulation converts robot control into a language modeling problem, enabling the direct reuse of LLM training and inference mechanisms. However, it introduces a discretization layer that approximates continuous dynamics and may impact control precision. OpenVLA adopts a simple yet scalable approach: fine-tuning a pretrained VLM to map images and instructions to action tokens \cite{openvla, openx}. This avoids task-specific architectural design and leverages large-scale datasets such as Open X-Embodiment \cite{openx} for training. This design emphasizes the scalability and reuse of existing models over fine-grained control optimization. It aligns with a broader trend in robotics toward foundation models, where generality is prioritized over task-specific performance.

\subsection{OpenVLA-OFT}
Although OpenVLA provides a clear and unified formulation of VLA models, it also exposes practical limitations when considered for real-world deployment, particularly in terms of inference speed, action representation, and control frequency. Building on these observations, OpenVLA-OFT (Optimized Fine-Tuning) was introduced as a refinement of the original model, aiming to revisit key design choices while preserving the overall architecture \cite{openvla_oft}. The OpenVLA-OFT work identifies three major limitations in the original OpenVLA formulation. First, the autoregressive decoding scheme requires sequential prediction of action tokens. At each timestep, the model generates the seven action components (position, orientation, gripper) one after another, resulting in significant latency and limiting control frequency. This sequential dependency also prevents efficient prediction of multi-step trajectories, which is particularly limiting in tasks requiring long-horizon planning or where predicting several steps ahead would allow smoother and more reactive control. Second, the discrete action representation, based on 256-bin quantization, introduces an approximation layer between the model output and the continuous control space. This discretization can limit precision and introduces additional complexity in tokenization and de-tokenization. Third, the training objective, formulated as next-token prediction with cross-entropy loss, is inherited from language modeling and may not be optimally aligned with continuous control objectives.

\begin{figure}[H]
    \centering    
    \includegraphics[width=12cm]{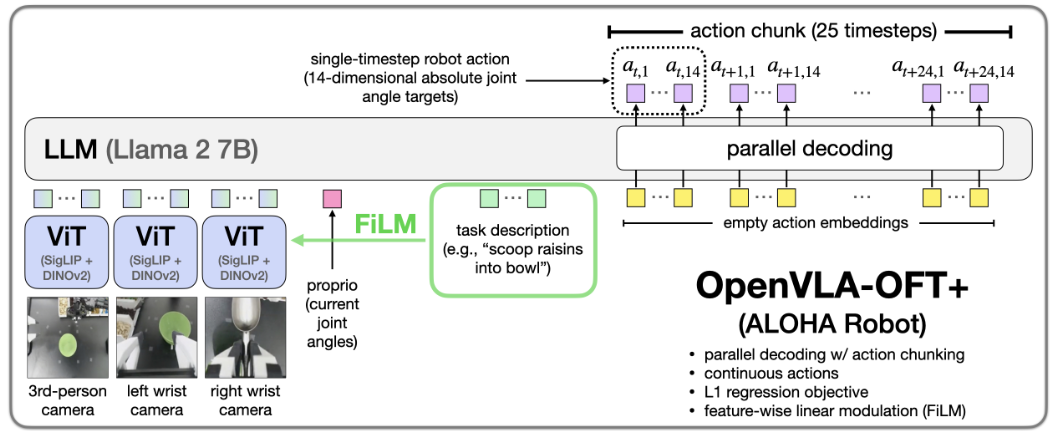}
    \caption{OpenVLA-OFT architecture from \cite{openvla_oft}}
    \label{fig:openvla-oft-arch}
\end{figure}

To address these limitations, OpenVLA-OFT introduces a set of coordinated modifications that target action generation, representation, and learning objectives. 
\textit{(i) Parallel action decoding} The autoregressive decoding scheme is replaced by a parallel formulation. Instead of generating each action dimension sequentially, the model predicts all action components simultaneously in a single forward pass using bidirectional attention. This modification removes the sequential dependency and significantly reduces inference latency. More importantly, it enables a more direct mapping between latent representations and full control commands.
\textit{(ii) Action chunking} Parallel decoding naturally enables action chunking, where the model predicts multiple future timesteps at once. For a chunk size (K), the model outputs a sequence of (K) actions in a single pass, which can then be executed open-loop. This approach improves both computational efficiency and trajectory consistency, as it allows the model to reason over short temporal horizons rather than isolated timesteps.
\textit{(iii) Continuous action representation} OpenVLA-OFT replaces discrete action tokens with a continuous action representation. Instead of mapping outputs to token probabilities, the hidden states of the decoder are passed through a dedicated linear (MLP) head that directly predicts normalized continuous actions. This removes the need for discretization and enables a more faithful representation of robot control signals.
\textit{(iv) Regression-based learning objective} Consistent with the continuous representation, the training objective is reformulated as an L1 regression loss over actions, replacing the cross-entropy loss used in OpenVLA. This change aligns the optimization objective with the underlying control problem and has been shown to provide competitive performance while improving convergence speed.
\textit{(v) Extended input modalities} In addition to the third-person camera used in OpenVLA, OpenVLA-OFT incorporates wrist-mounted camera observations and optionally robot state (proprioception) as inputs. This multi-view setup provides richer spatial information, particularly for manipulation tasks involving occlusions or fine-grained interactions.
The OpenVLA-OFT architecture is presented in Figure~\ref{fig:openvla-oft-arch}.

The experimental results reported by the authors indicate that these modifications lead to consistent improvements across multiple dimensions \cite{openvla_oft}. In particular, OpenVLA-OFT achieves significantly higher inference efficiency, with reported speedups up to 26× through parallel decoding and action chunking, faster training convergence under the L1 regression objective, and improved task success rates on both simulated benchmarks and real-world robotic tasks .

Taken together, these results highlight a critical insight: the limitations observed in OpenVLA are not solely due to model capacity but are strongly influenced by the formulation of the action interface and decoding strategy. By moving from a language-inspired formulation (discrete tokens, autoregressive decoding) to a control-oriented formulation (continuous actions, parallel prediction, regression loss), OpenVLA-OFT bridges part of the gap between multimodal reasoning models and real-time robotic control.

\subsection{Evaluation Paradigms for VLA Models in Simulation and Real-World Settings}
Evaluation of VLA models remains an evolving and partially unresolved challenge, particularly when transitioning from controlled benchmarks to real-world robotic systems. Existing evaluation paradigms are largely rooted in simulation-based or semi-controlled environments, where task definitions, observation spaces, and success metrics can be standardized. Although these frameworks have enabled rapid progress and comparative analysis between models, they only partially capture the constraints and variability inherent to physical deployment \cite{vla_survey, openvla}. A significant portion of the current evaluation landscape relies on structured benchmarks such as LIBERO \cite{libero_2023} and CALVIN \cite{calvin_2022}, which provide standardized task suites for assessing generalization, multi-task learning, and compositional reasoning in manipulation settings. LIBERO, in particular, introduces a family of task distributions designed to evaluate cross-task generalization under varying degrees of distribution shift while maintaining a consistent robotic embodiment. Similarly, CALVIN focuses on long-horizon sequential manipulation tasks with language conditioning, emphasizing the ability of policies to chain skills over time. These benchmarks have contributed to the establishment of common evaluation protocols, including success rates over predefined tasks, rollout-based metrics, and variations in initial conditions.

However, recent extensions of LIBERO, notably LIBERO-PRO \cite{zhou2026liberoprorobustfairevaluation} and LIBERO-PLUS \cite{fei2025liberoplusindepthrobustnessanalysis}, highlight emerging limitations in current benchmarking practices. These works emphasize that even within simulated or semi-controlled environments, evaluation results can be sensitive to factors such as task design, distributional assumptions, and the degree of alignment between training and evaluation conditions. In particular, LIBERO-PRO introduces more challenging and realistic task compositions, exposing limitations in policy robustness and generalization that are not apparent in simpler benchmark settings. LIBERO-PLUS further explores the scalability of evaluation protocols and the impact of increased task diversity, suggesting that standard success metrics may not fully capture nuanced failure modes or partial task completion. Importantly, these works do not invalidate existing benchmarks, but rather refine the understanding of what they measure, highlighting the need for more comprehensive and discriminative evaluation criteria.

Despite these advances, a fundamental limitation persists: benchmark-based evaluation remains intrinsically constrained by the assumptions of the simulated or controlled environment in which it is defined. In real-world settings, it is not feasible to reproduce such controlled conditions at scale. Variability in sensing (e.g. lighting, camera calibration), actuation (e.g. latency, control noise), and environment (e.g. object properties, workspace geometry) introduces a level of stochasticity that is difficult to standardize. As a result, defining reproducible and comparable evaluation protocols for physical systems becomes inherently challenging. Unlike simulation benchmarks, where identical initial states can be enforced, real-world experiments are subject to irreducible variability, making statistical comparisons across runs more complex and often less reliable.

In the context of the present work, the evaluation protocol differs from benchmark-based approaches such as LIBERO and CALVIN in that it relies on real-world closed-loop execution rather than standardized simulated rollouts. As a result, the evaluation focuses primarily on qualitative trajectory behavior, stability under recursive execution, and sensitivity to environmental variations, rather than aggregated success metrics. This distinction is important because it limits direct comparability with existing benchmarks while providing complementary insights into deployment-specific failure modes.

\section{Experimental Setup}

\subsection{Hardware}
This section describes the physical setup used for data collection and real-robot experiments, including the manipulation platform, sensing modalities, and communication interfaces. The robotic platform is built around a UR5e collaborative manipulator, equipped with its standard controller and a teach pendant for manual operation and safety supervision. The end-effector consists of a Robotiq 2F-140 parallel gripper, allowing basic grasping and manipulation tasks. The perception system relies on a dual-camera configuration: a third-person (“third-eye”) camera (Intel RealSense D435), which provides a global view of the workspace, and a wrist-mounted (“hand-eye”) camera (Logitech webcam), that offers a close-range perspective aligned with the end-effector. Robot control and data acquisition are performed from an external computer using the $ur\_rtde$\footnote{\url{https://pypi.org/project/ur-rtde/}} Python library, which provides the RTDE Control Interface to send motion commands, the RTDE Receive Interface to access robot state feedback, and an additional API to control the Robotiq gripper.

\begin{figure}[H]
    \centering    
    \includegraphics[width=8cm]{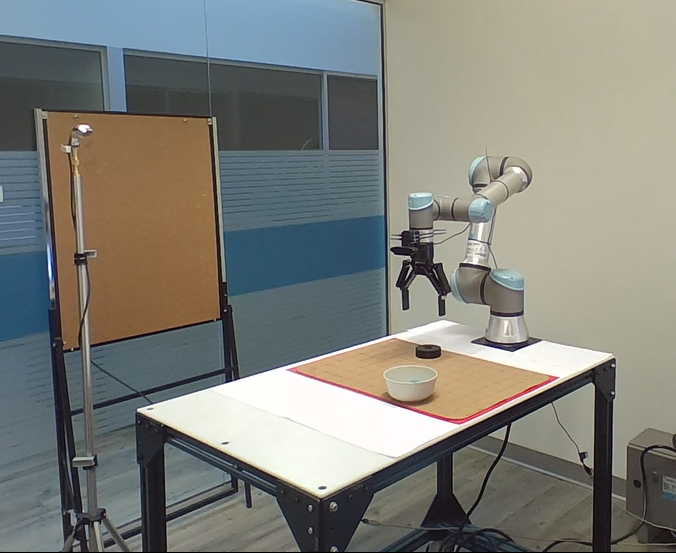}
    \caption{Experimental robotic plateform used for the project (UR5e, Gripper Robotic 2F-140, Vention supports, third-eye IntelRealSense D435 camera and hand-eye Logitech Webcam)}
    \label{fig:hardware}
\end{figure}

This hardware configuration reflects a standard, yet representative, setup for manipulation tasks with VLA models. The combination of a fixed external camera and an egocentric wrist camera enables complementary visual information: global scene understanding and local interaction details. The use of the RTDE interface allows for low-level, programmatic control of the robot while maintaining compatibility with real-time constraints. At the same time, the teach pendant facilitates manual data collection (e.g., demonstrations via freedrive), which is critical for imitation learning pipelines.

Overall, the selected hardware provides a practical balance between simplicity, flexibility, and relevance to real-world manipulation scenarios (e.g. in manufacturing context). It supports both data acquisition and closed-loop inference, while remaining sufficiently modular to accommodate future extensions (e.g., additional sensors or control interfaces).

\subsection{Software Environment}
This section describes the software architecture and execution environment used for model inference, data processing, and fine-tuning.

The experimental setup follows a client–server architecture. The client runs on the local machine that controls the robot, while the server is deployed on a compute node within an on-prem cluster. This design is motivated by the computational requirements of VLA models, which typically rely on high-performance GPUs with large memory capacity (e.g. NVIDIA A100), not available on the robot control workstation. In this configuration, sensory inputs (camera images) are acquired on the client side and transmitted to the server for inference. The predicted actions are then returned to the client and executed on the robot. Although this introduces additional data transfer steps, the impact on latency remains limited in practice. The observed inference frequency of the tested models is approximately 3 Hz\footnote{Measured using the original OpenVLA on an A100 GPU}, making network transfer a secondary contributor to overall system latency. To simplify integration, the model is wrapped as a lightweight Gradio-based\footnote{\url{https://www.gradio.app/}} application on the server side. This provides a simple API for sending image input and receiving action predictions, reducing development overhead, and enabling rapid experimentation.

The software environment is Python-based and managed using conda environment. 
\textit{Server side}: the environment provided by the OpenVLA authors is used with minimal modifications, ensuring compatibility with the original training and inference pipelines.
\textit{Client side}: the environment includes the $ur\_rtde$\footnote{\url{https://pypi.org/project/ur-rtde/}} library for robot control and state feedback. In terms of hardware support, local experimentation and simulation are performed on a workstation equipped with an NVIDIA GeForce RTX 5080 (16 GB) running Ubuntu 22.04, and the model fine-tuning is conducted on a cluster using 4× NVIDIA A100 GPUs. This architecture reflects a trade-off between computational requirements and system simplicity. Offloading inference to a remote server enables the use of large models without modifying the robot-side infrastructure. At the same time, the relatively low inference frequency of current VLA models reduces the impact of network-induced latency, making this approach viable for experimental purposes. The use of a Gradio wrapper further emphasizes a pragmatic approach: prioritizing rapid prototyping and ease of integration over optimized deployment. This is consistent with the project’s objective of understanding the behavior of the model rather than maximizing real-time performance.

\section{Methodology}

\subsection{Global Overview of the Applied Deployment Workflow}
This section provides a structured overview of the experimental workflow implemented throughout the project. The objective is to progressively evaluate VLA models under increasing levels of realism, from controlled offline validation to real-world deployment.

\begin{figure}[H]
    \centering    
    \includegraphics[width=12cm]{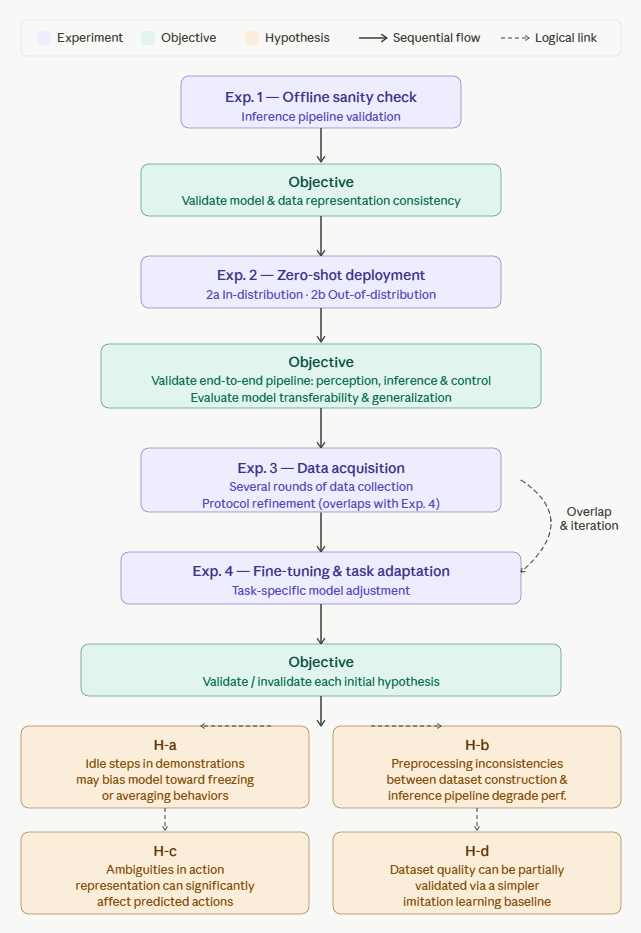}
    \caption{Global overview of the applied deployment workflow.}
    \label{fig:workflow}
\end{figure}

This section provides a structured overview of the experimental process implemented throughout the project. The goal is to progressively evaluate the VLA models at increasing levels of realism, from controlled offline validation to deployment under real-world conditions. The experimental process consists of four stages: 1) An offline validation experiment using examples from the training dataset. The goal of this stage is to validate the overall consistency of the inference chain under controlled conditions. 2) A zero-shot experiment on episodes both within and outside the training dataset. This stage consists of evaluating the model’s ability to adapt to a real robotic system without fine-tuning. 3) \& 4) Data acquisition and model fine-tuning. The two experiments are linked and are conducted iteratively. These experiments involve adapting the model to the local environment through the collection of specific data. These four experiments, conducted on the order of 1 to 3/4, aim to validate the four research hypotheses defined in Section 2.2. Research questions and hypothesis. As a reminder, the first hypothesis is that the presence of idle steps in demonstrations may bias the model toward freezing or averaging behaviors, rather than producing goal-directed action. The second is that inconsistencies between image preprocessing during dataset construction and the internal preprocessing pipeline used during training and inference are expected to degrade performance. The third one addresses the fact that ambiguities in action representation, particularly regarding reference frames and normalization/de-normalization conventions, are hypothesized to significantly affect predicted actions, limiting generalization across setup. Finally, the fourth hypothesis  suggests that the consistency and quality of the dataset and pipeline can be partially validated using a simpler imitation learning baseline. Figure~\ref{fig:workflow} presents a schematic illustration of the experiments carried out and the four hypothesis. 

\textit{1. Offline sanity check (in-distribution validation) : }
The first stage focuses on validating the high-level consistency of the inference pipeline under controlled conditions. A subset of episodes from an existing training dataset (Berkeley UR5\footnote{\url{https://sites.google.com/view/berkeley-ur5/home}}) is selected, specifically tasks that are reproducible with our own setup/material and involve simple object interactions (e.g., cloth sweeping). Using a pretrained OpenVLA model in its out-of-the-box configuration, inference is performed step by step by providing the model with the original training input (image and instruction) at each timestep.The predicted actions are then compared to the ground-truth actions (these ground-truth actions are calculated from the actual robot states measured at each time step during the episode recording and are directly accessible in the Berkeley documentation). This setup enables closed-loop evaluation without introducing a distribution offset. This setup allows evaluation in a closed-loop manner, without introducing a distribution shift. The objective is to verify that the model, data format, and inference pipeline are aligned correctly in an offline setting.

\textit{2. Zero-shot deployment on the real robot}
The second stage evaluates the model’s ability to generalize to a real robotic system without fine-tuning. Two scenarios are considered. (i) In-distribution approximation: the experimental setup is configured to closely match the original dataset conditions (e.g., camera placement, task structure), in order to minimize distribution shift. (ii) Out-of-distribution evaluation : the model is tested on a different task (e.g., pick-and-place), introducing variations in both task definition and visual context.Inference is performed using live camera input from the real system, enabling complete online closed-loop execution. The objective of this stage is to validate the end-to-end deployment pipeline, including perception, inference, and control, under real-world conditions.

\textit{3. Fine-tuning and task-specific adaptation}
The third stage focuses on adapting the model to the local environment through task-specific data collection and fine-tuning. This process follows a structured pipeline: 1) definition of the task and workspace, 2) design of a data acquisition protocol, 3) scripted data collection, 4) preprocessing and dataset conversion, 5) model fine-tuning using LoRA and 6) validation of the collected data through replay of recorded trajectories. Following fine-tuning, the model is then evaluated in multiple configurations: offline inference, simulation (open-loop and closed-loop), and real-robot inference using both recorded and live inputs. This progressive evaluation ensures safety while enabling the analysis of model behavior across different levels of deployment.

This staged workflow reflects a deliberate strategy to isolate and evaluate different sources of failure. The initial offline validation ensures that the model and data representations are consistent. Zero-shot experiments introduce real-world variability and test generalization. Finally, the fine-tuning stage assesses whether task-specific data can bridge the gap between model capabilities and deployment requirements. By structuring the experiments in this way, the workflow enables systematic investigation of the central hypothesis introduced earlier: that real-world performance is governed by the alignment of the full data–model–control pipeline, rather than by model capacity alone.

The deployment workflow provides a reproducible framework for evaluating VLA models across multiple levels of complexity. It supports both controlled validation and real-world testing while enabling detailed analysis of failure modes at each stage. This structured approach is essential to understand the limitations of current VLA systems and guide future improvements in data collection, model adaptation, and system integration.

\subsection{Acquisition Protocol and Dataset}

This section describes the data acquisition strategies and dataset construction process used for model training and evaluation. Two main acquisition strategies are implemented: scripted acquisition and freedrive (manual) acquisition, both relying on the same hardware setup described previously.

\subsubsection{Scripted acquisition}
\begin{figure}[H]
    \centering    
    \includegraphics[width=12cm]{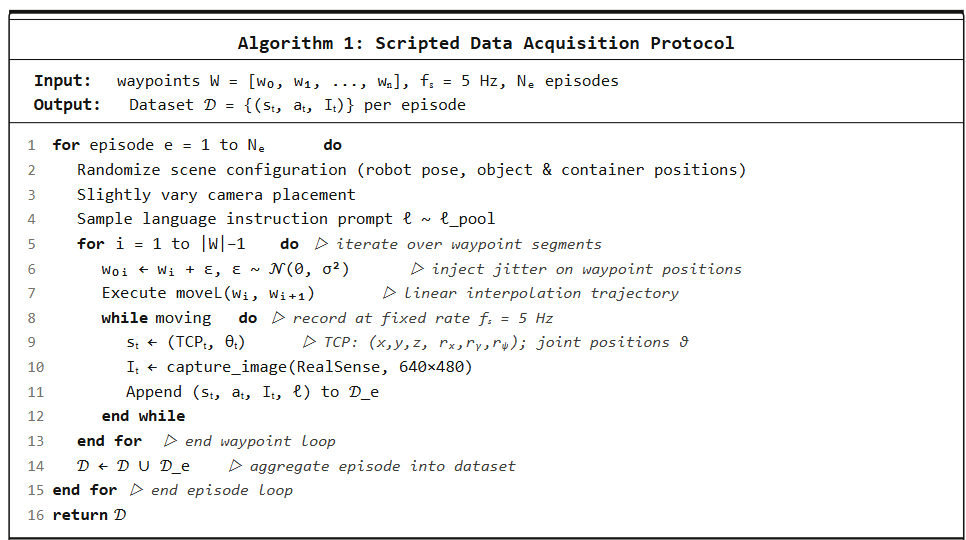}
    \caption{Scripted protocol. Waypoint-based trajectories with injected variability. Variability introduced through scene randomisation, prompt sampling, and waypoint jitter.}
    \label{fig:algorithm1}
\end{figure}

In the scripted setting, each task is defined through a sequence of manually specified waypoints\footnote{A waypoint is a designated geographical coordinate used to define a route, navigation path, or location.} corresponding to key stages of execution. For example, in a pick-and-place task, the trajectory typically includes: an initial home position, a pre-grasp approach pose, a grasping pose at the object location, a transport phase, and a placement pose at the target location. The robot executes trajectories between these waypoints using linear interpolation (e.g., $moveL$ commands). Data is recorded at a fixed frequency of 5 Hz, capturing at each timestep: the Cartesian position of the tool center point (TCP) $(x, y, z)$ and the orientation as a rotation vector ($r_x$, $r_y$, $r_z$) in the base frame (the fixed reference frame attached to the robot's mounting point, as opposed to the tool frame which moves with the end-effector), joint positions, and RGB images from the third-person camera (Intel RealSense with $640*480$ pixel resolution). The main logic of the code used to collect scripted trajectories is presented in Figure~\ref{fig:algorithm1}. To avoid overfitting to deterministic trajectories, variability is introduced across episodes by modifying the initial scene configuration (robot pose, object positions, container locations), slightly varying camera placement, changing the language instruction (prompt), adding small perturbations (jitter) to waypoint positions. The robot motion is executed with approximately constant velocity and acceleration, with minor variations due to injected noise. This approach enables rapid acquisition of large volumes of structured data, but relies on predefined trajectories and limited interaction variability.

\subsubsection{Freedrive acquisition}
\begin{figure}[H]
    \centering    
    \includegraphics[width=12cm]{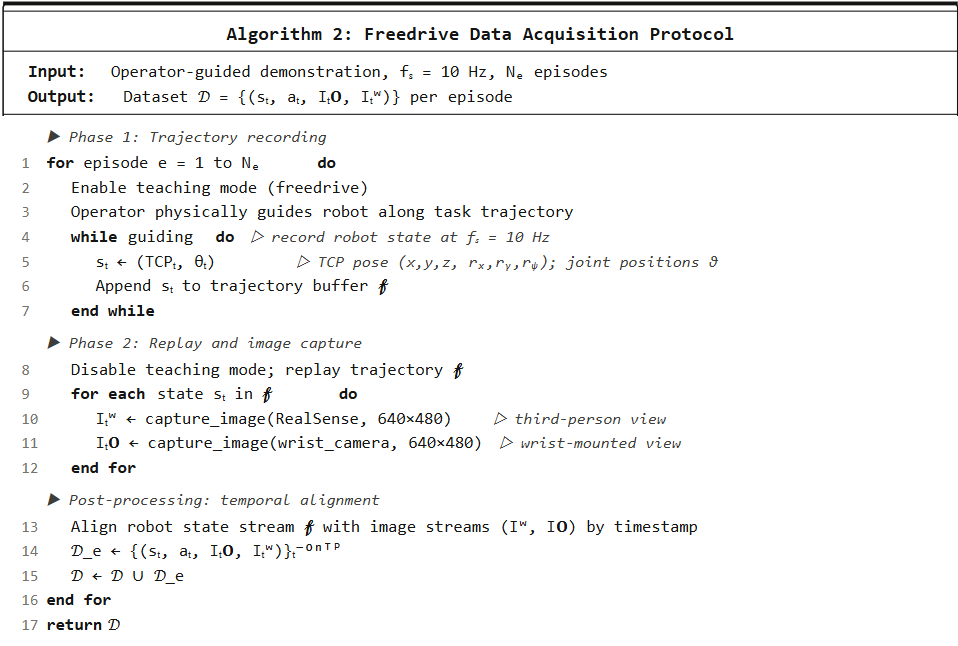}
    \caption{Freedrive protocol. Two-phase collection: manual guidance records state at 10Hz followed by trajectory replay for image capture with post-hoc temporal alignment.}
    \label{fig:algorithm2}
\end{figure}

In the freedrive setting, demonstrations are collected by physically guiding the robot using its teaching mode. An operator manually moves the robot along the desired trajectory, providing more natural and task-adaptive motions. Data acquisition is performed in two stages: 1) Trajectory recording : the robot state (TCP pose and orientation and joint positions) is recorded at 10 Hz during manual guidance. 2) Replay and image capture: the recorded trajectory is replayed and images are captured from both the third-person and wrist-mounted cameras. A post-processing step is then required to temporally align the robot states and image streams. This approach captures more realistic motion patterns and implicit human strategies, but is more time-consuming and introduces additional processing complexity. The main logic of the code used for freedrive acquisition is presented in Figure~\ref{fig:algorithm2}.

\subsubsection{Dataset structure}
This section describes the structure of the datasets constructed for fine-tuning and evaluation, as well as the progressive increase in data complexity across dataset versions (presented in Table~\ref{tab:dataset_versions}).

\begin{table}[H]
    \centering
    \caption{Summary of dataset versions used for fine-tuning and evaluation}
    \label{tab:dataset_versions}
    \begin{tabular}{p{1cm} p{1cm} p{1cm} p{2.5cm} p{3cm} p{4.5cm}}
        \toprule
        Version & Episodes & Mode & Mean Episode Length (freq) & Task / Instruction & Description \\
        \midrule
        V0 & 100 & Scripted (no noise) & 60 frames per ep. (5Hz) & Pick up the apple and place it in the bowl & Fixed third-person camera. Variation in initial robot pose and object positions. \\
        V1 & 95 & Scripted (with noise) & 120 frames per ep. (10Hz) & Put apple into {pot, big bowl}; lift {white, black} cup & Fixed camera. Variation in object positions only. Addition of noise in trajectories. \\
        V2 & 30 & Scripted (no noise) & 60 frames per ep. (5Hz) & Pick and place {orange, apple}; pick {apple, marker}; put apple on plate & Fixed camera. Variation in robot and object initial poses. Introduction of distractor elements. \\
        V3 & 40 & Scripted (with noise) &  from  120 to 300 frames per ep. (10Hz) & Pick up screwdriver and place in box; multi-object placement task & Variable camera position. Increased task complexity and scene variability. \\
        V4 & 120 & Scripted (with noise) & 120 frames per ep. (10Hz) & Multiple phrasings of gear pick-and-place task & Variable camera position. Variation in object positions and language instructions. Addition of distractors. \\
        V5 & 30 & Freedrive (manual) & 200 frames per ep. (10Hz) & Put apple into pot & Fixed camera. Variation in object positions. Human-guided trajectories. \\
        \bottomrule
    \end{tabular}
\end{table}

Across these versions, several axes of variation are progressively introduced: scene variability (object positions, initial robot pose), trajectory variability (noise vs. deterministic motion), task diversity (single task vs. multiple tasks), language variability (multiple phrasings for similar tasks), viewpoint variation (fixed vs. moving camera), and acquisition mode (scripted vs. human-guided (freedrive)).
The dataset design follows an incremental strategy, where complexity is gradually increased to assess the sensitivity of VLA models to different sources of variation. The early datasets (V0–V1) focus on controlled conditions with limited variability, enabling isolation of basic behaviors. Subsequent versions (V2–V4) introduce increased diversity in tasks, scenes, and viewpoints, while V5 introduces a shift in acquisition modality toward more realistic human-guided trajectories. This progression is intended to investigate how different aspects of the data (such as visual diversity, language variation, and trajectory structure) affect model learning and generalization. In particular, it enables controlled comparison between highly structured datasets and more natural but less regular demonstrations.

\begin{figure}[htbp]
    \centering
    \begin{subfigure}[b]{0.48\textwidth}
        \centering
        \includegraphics[width=\textwidth]{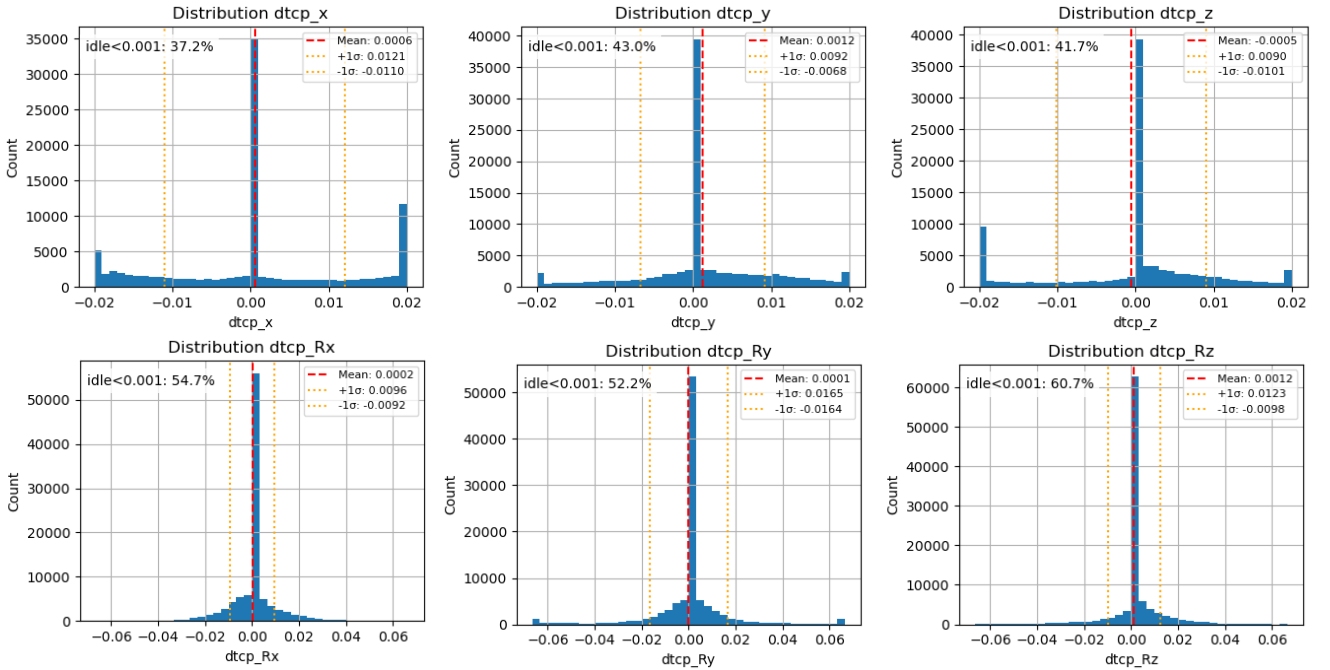}
        \caption{Berkeley UR5 (reference, y-axis: count)}
    \end{subfigure}
    \hfill
    \begin{subfigure}[b]{0.48\textwidth}
        \centering
        \includegraphics[width=\textwidth]{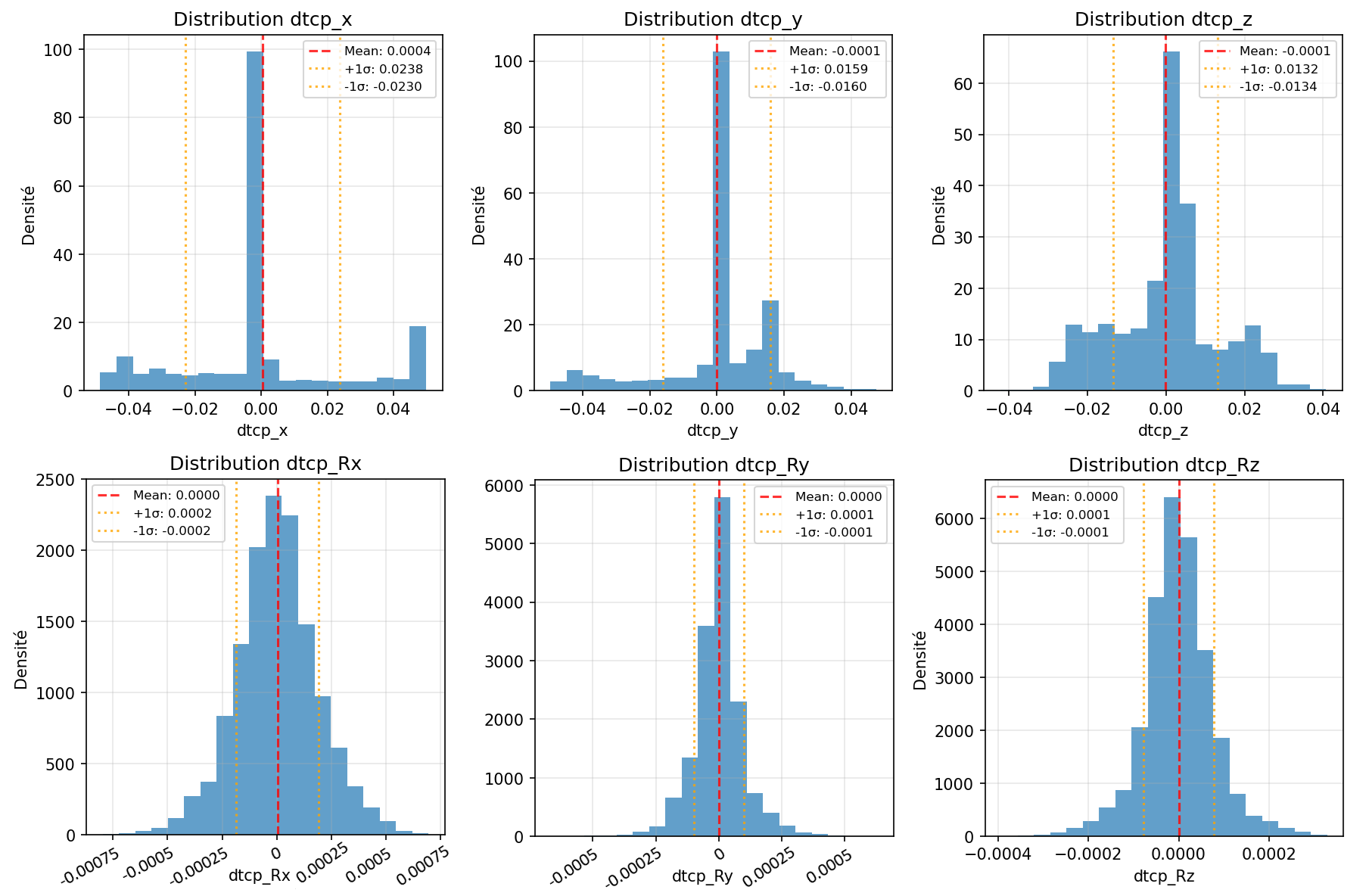}
        \caption{Dataset V0 (scripted, no noise, y-axis: density)}
    \end{subfigure}

    \vspace{0.5em}

    \begin{subfigure}[b]{0.48\textwidth}
        \centering
        \includegraphics[width=\textwidth]{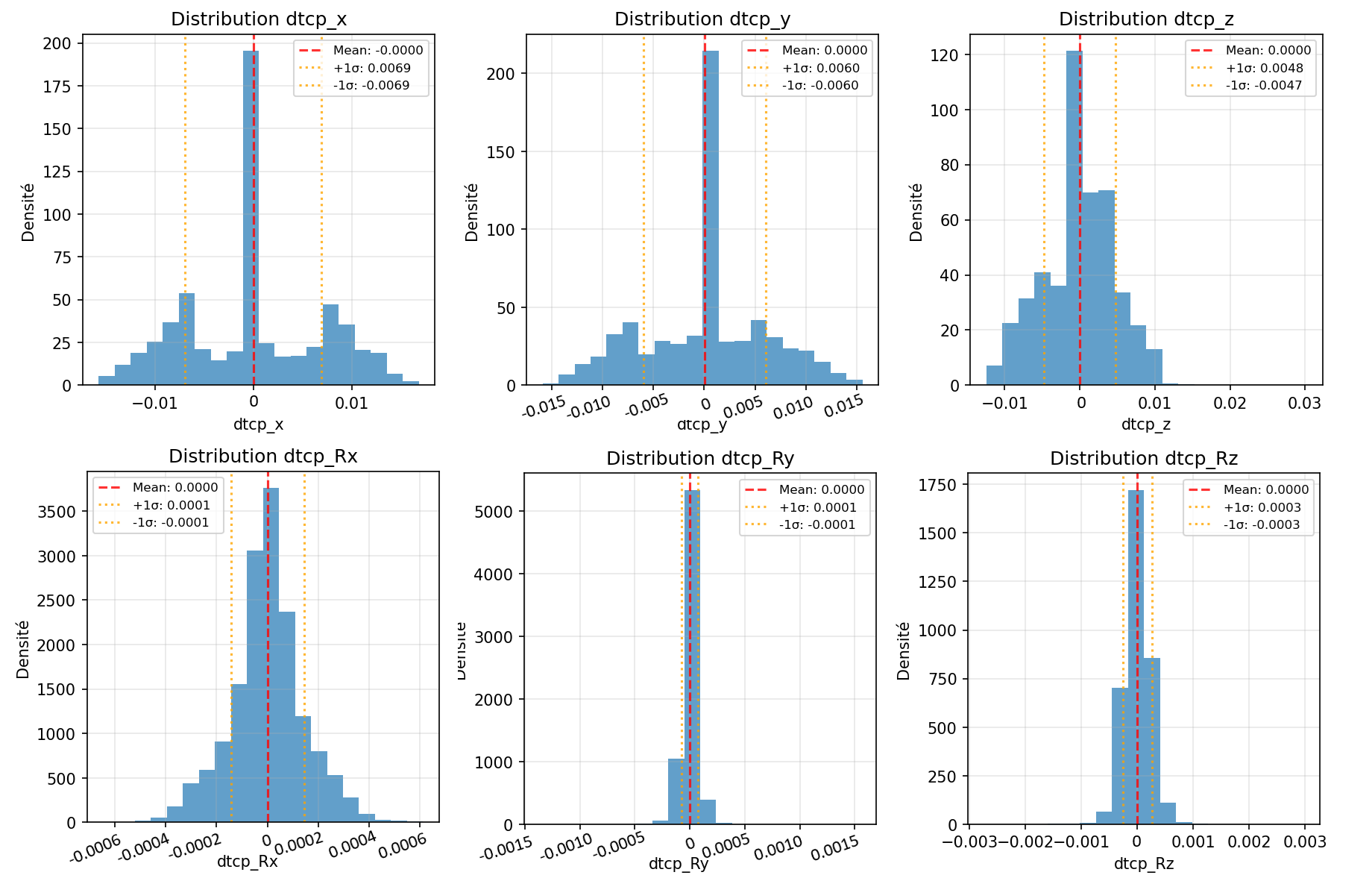}
        \caption{Dataset V4 (scripted, with noise and variability, y-axis: density)}
    \end{subfigure}
    \hfill
    \begin{subfigure}[b]{0.48\textwidth}
        \centering
        \includegraphics[width=\textwidth]{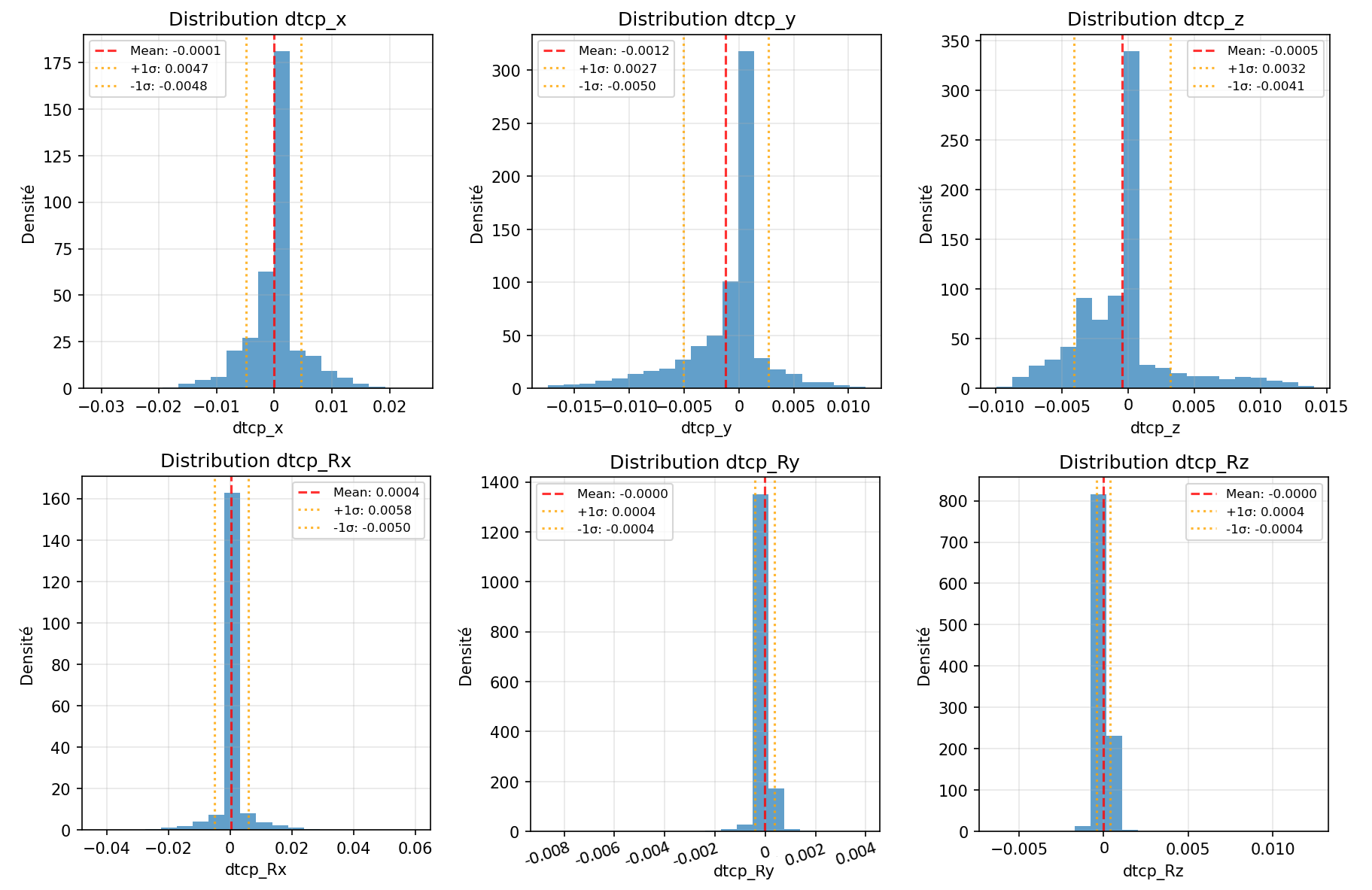}
        \caption{Dataset V5 (freedrive, human-guided, y-axis: density)}
    \end{subfigure}

    \caption{Action distributions $(\Delta x, \Delta y, \Delta z, 
    \Delta r_x, \Delta r_y, \Delta r_z)$ across selected dataset 
    versions compared to the Berkeley UR5 reference. Note that 
    Berkeley is displayed in raw counts while local datasets are 
    displayed in density, reflecting differences in dataset scale; 
    the comparison focuses on distribution shape rather than absolute 
    values. The gripper dimension is excluded from this comparison.}
    \label{fig:action_distributions}
\end{figure}

Dataset construction followed the best practices recommended in the official OpenVLA repository \cite{openvla_repo}, namely a control frequency in the 5–10 Hz range and minimization of idle steps. To verify alignment with the reference distribution, the action statistics were visually compared with the Berkeley UR5 dataset across all Cartesian action dimensions $(\Delta x, \Delta y, \Delta z, \Delta r_x, \Delta r_y, \Delta r_z)$, as shown in Figure \ref{fig:action_distributions} (note that the Berkeley histogram is displayed in raw counts while local datasets use density normalization, reflecting differences in dataset scale; the comparison therefore focuses on distribution shape.). Interpretation of these distributions requires accounting for the acquisition frequency and trajectory structure. Datasets collected at 5 Hz (V0) produce larger per-step deltas than those at 10 Hz (V4, V5), resulting in wider marginal distributions that do not necessarily reflect greater task diversity. In fact, V0 is the most deterministic dataset: scripted without noise injection, its wide but sharply peaked distributions reflect a small set of recurring, stereotyped motions rather than genuine variability. A more informative indicator of diversity is therefore the flatness of the distribution: later versions exhibit lower kurtosis across Cartesian dimensions, reflecting increased directional diversity despite smaller step amplitudes. This progression confirms the intended design of the dataset versioning strategy, where diversity is gradually introduced through noise injection, task variation, and ultimately human-guided demonstrations.

While this chronological dataset design enables controlled exploration of variability factors, the experiments indicate that improvements in dataset diversity do not directly translate into stable closed-loop behavior. In particular, the datasets remain dominated by successful, scripted trajectories with limited representation of recovery behaviors or off-distribution states. This may contribute to the observed inability of the models to correct deviations during execution, highlighting a limitation in distribution coverage rather than dataset size alone.
\subsection{Data Processing}
This section describes the transformations applied to the raw data after acquisition, with the objective of producing a consistent dataset compatible with VLA training pipelines. Following acquisition, all data are standardized and converted into a format aligned with the requirements of OpenVLA and, more broadly, with datasets derived from the Open X-Embodiment framework. This implies structuring the data according to the RLDS (Reinforcement Learning Datasets) format \cite{rlds}, a standard designed for sequential decision-making data. RLDS organizes data hierarchically. Episodes represent complete task executions, and each episode is composed of steps. Each step contains: an observation (e.g., images, robot state), an action, optional signals such as reward or discount, and metadata flags (e.g., $is\_first$, $is\_last$, $is\_terminal$).

The preprocessing pipeline consists of the following steps. \textit{(1) Action computation :} Actions are derived from recorded robot states and correspond to the quantities that the VLA model should predict (ground truth). For each timestep, actions are computed as temporal differences between consecutive states:  $(\Delta x, \Delta y, \Delta z, \Delta r_x, \Delta r_y, \Delta r_z, \Delta g)$ where $((x, y, z, r_x, r_y, r_z))$ denotes the TCP pose (in base frame) and $(\Delta g)$ represents the gripper command. \textit{(2) Gripper discretization :} The gripper signal is converted into a binary open/close command using a manually defined threshold. \textit{(3) Image preprocessing :} Depending on the experiment, the images can be resized, cropped, or adjusted to match the expected input format of the model. \textit{(4) Temporal synchronization and structuring :} Robot states, actions, and image streams are temporally aligned. The data is then organized into episode-wise arrays, where each timestep corresponds to a single sample containing: static camera image, wrist camera image, robot state $(x, y, z, r_x, r_y, r_z, q_1, \dots, q_6, g)$, action $(\Delta x, \Delta y, \Delta z, \Delta r_x, \Delta r_y, \Delta r_z, \Delta g)$, and episode metadata. \textit{(5)Conversion to RLDS format :} The structured data are finally converted to RLDS format using an open-source dataset builder\footnote{\url{https://github.com/kpertsch/rlds_dataset_builder}}, ensuring compatibility with existing VLA training pipelines. The conversion to RLDS ensures interoperability with existing datasets and models, but imposes strict requirements on data structure and consistency. Any mismatch between preprocessing and model expectations (e.g., image transformations or action normalization) can directly impact performance.

\subsection{Fine-Tuning Workflow}
This section describes the practical fine-tuning procedures used for OpenVLA and OpenVLA-OFT. The objective is to present an operational view of the training pipeline, highlighting key configuration choices and implementation constraints.

\subsubsection{OpenVLA}
The OpenVLA fine-tuning procedure follows the guidelines provided by the maintainers of the official repository \cite{openvla_repo}. Due to the high computational requirements of full model training (e.g., multi-node setups with up to 8× A100 GPUs), the parameter-efficient fine-tuning approach based on LoRA \cite{lora_2021} is chosen. The training process is implemented using the fine-tuning script provided by the authors, which defines a \textbf{$FinetuneConfig$} class encapsulating the main hyperparameters. The default configuration is presented in Table~\ref{tab:params_finetuning_openvla}.

\begin{table}[]
    \centering
    \begin{tabular}{|c|c|c|}
    \hline
       Parameter  & Default value  \\
       \hline
       vla\_path & openvla/openvla-7b \\
       batch\_size  & 16  \\
       max\_steps  & 200k   \\
       learning\_rate  & 5e-4  \\
       image\_aug  & True  \\
       use\_lora & True \\
       \hline
    \end{tabular}
    \vspace{5pt}
    \caption{Default training parameters for OpenVLA LoRA fine-tuning}
    \label{tab:params_finetuning_openvla}
\end{table}

In addition to these defaults, two parameters must be explicitly specified: $data\_root\_dir$: path to the dataset in RLDS format, and  $dataset\_name$: identifier used to load both the dataset and its associated preprocessing configuration\footnote{\url{https://github.com/openvla/openvla/tree/main?tab=readme-ov-file\#fine-tuning-openvla-via-lora}}. 
A notable design choice made by the authors in this workflow is the absence of a validation loop in the training script. According to the authors, validation accuracy is not a reliable predictor of downstream task success\footnote{\url{https://github.com/openvla/openvla/issues/33}} and therefore is not included in the default training procedure.

\subsubsection{OpenVLA-OFT}
The fine-tuning procedure for OpenVLA-OFT is derived from the same codebase, as the project is in fact a fork of the original OpenVLA code repository. As a result, the same training script has been enhanced with additional configuration options reflecting the architectural differences introduced by OFT. In particular, the OFT pipeline introduces additional training components and configuration choices, increasing the overall complexity of the fine-tuning process. One such component is the FiLM (Feature-wise Linear Modulation) module, which enhances image grounding: This mechanism is intended to encourage the model to better condition its predictions on language inputs. Although presented as optional, the use of FiLM appears to be task- and setup-dependent. In multi-view configurations (e.g., with both third-person and wrist-mounted cameras), it is described as essential to prevent overfitting to spurious visual features and to ensure proper alignment between visual and linguistic information. However, its impact may vary depending on the dataset, making it a parameter that requires empirical validation.
Another distinction with OpenVLA is the presence of a validation loop in the OFT training pipeline, enabling monitoring of training dynamics through intermediate evaluation metrics.

The fine-tuning workflows highlight a key tension between practical constraints and methodological rigor. On the one hand, the use of LoRA enables efficient adaptation of large models under limited computational resources. On the other hand, the lack of validation metrics, particularly in the OpenVLA pipeline, complicates model selection and performance assessment. Furthermore, the increased flexibility of the OFT pipeline introduces additional degrees of freedom, which may improve performance but also increases sensitivity to configuration choices and inevitably increases the exploration space. The fine-tuning process is not a fully standardized procedure but rather a configuration-dependent pipeline requiring careful alignment between dataset structure, model assumptions, and training parameters. This reinforces the broader observation that successful deployment of VLA systems depends not only on model architecture, but also on the precise control of the training workflow within the data–model–control pipeline.

\section{Experiments and results}

This section reports the main experimental observations obtained throughout the project, following the staged workflow introduced previously. The results are organized according to four main evaluation settings: offline sanity check, zero-shot deployment on the real robot, task-specific fine-tuning of OpenVLA and OpenVLA-OFT, and comparison with a simpler imitation-learning baseline.

\subsection{Offline Sanity Check}
\begin{figure}[htbp]
    \centering
    \begin{subfigure}[b]{0.45\textwidth}
        \centering
        \includegraphics[width=\textwidth]{{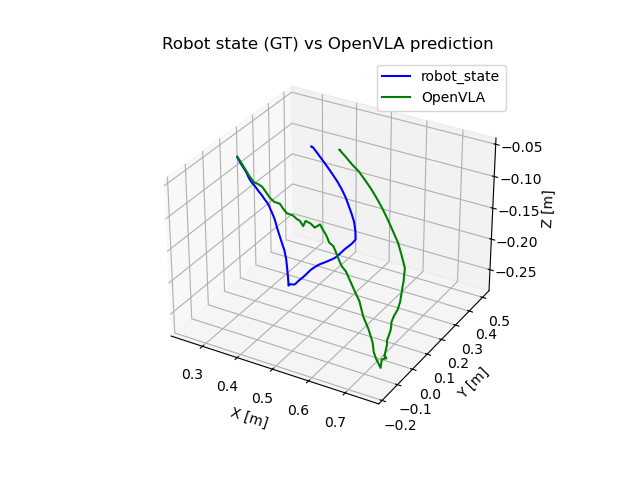}}
        \caption{GT vs OpenVLA predictions}
        \label{fig:sanity_check}
    \end{subfigure}
    \begin{subfigure}[b]{0.45\textwidth}
        \centering
        \includegraphics[width=\textwidth]{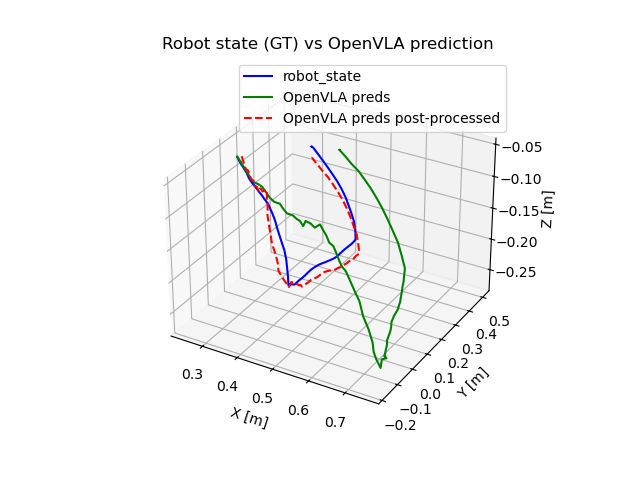}
        \caption{GT vs OpenVLA predictions with post-processing}
        \label{fig:postprocess}
    \end{subfigure}
    
    \caption{3D trajectory comparison (GT vs OpenVLA predictions). This figure shows the ground-truth trajectory together with OpenVLA predictions in 3D spaces, including raw (a) and post-processed (b) predictions.}
    \label{fig:berkeley}
\end{figure}

The first experiment aims to validate the consistency of the inference pipeline under controlled, approximately in-distribution conditions. This step serves as a preliminary verification before introducing real-world variability. A subset of episodes from the Berkeley UR5 dataset\footnote{\url{https://sites.google.com/view/berkeley-ur5/home}} is used for this evaluation. This dataset is part of the Open X-Embodiment \cite{openx} collection and represents approximately 1.2\% of the data used to train OpenVLA. It consists of manipulation tasks performed on a UR5 platform, making it particularly suitable for consistency checks with the current setup. Inference is performed step-by-step using a pretrained OpenVLA model in its out-of-the-box configuration\footnote{\url{https://huggingface.co/openvla/openvla-7b}}. At each timestep, the model receives the recorded image and language instruction from the dataset, and predicts the corresponding action. The predicted trajectories are then compared with the recorded ground-truth robot trajectories. At the trajectory level, the predicted motions follow the global shape of the ground-truth trajectories. The temporal evolution of the predicted Cartesian positions is also qualitatively consistent with the reference motion. 
However, noticeable discrepancies are observed in trajectory amplitude and spatial alignment when raw predictions are used directly.  To better characterize the nature of these discrepancies, two diagnostic post-processing steps are applied: a rigid-body alignment and a per-axis scale factor estimation. The rigid-body alignment consists of finding the optimal SE(3) (Spatial Euclidian) transformation (rotation and translation) that minimizes the distance between the predicted and ground-truth trajectories in the least-squares sense. In addition, a temporal lag parameter is estimated to account for potential phase shifts between the two signals. The per-axis scale factors are independently estimated along $x$, $y$, and $z$ to quantify the systematic compression of the amplitude introduced by the discrete action representation of the model. On the episode shown in Figure~\ref{fig:berkeley}, the raw predicted trajectory yields a root mean squared error (RMSE) of $0.1960m$. After SE(3) alignment, the RMSE drops further from $0.103m$ to $0.106m$ when a temporal lag of one timestep is additionally compensated. The estimated per-axis scale factors are $[0.635, 0.603, 0.722]$ along $(x,y,z)$  indicating a consistent underestimation of motion amplitude of approximately 30 to 40\%. These measurements suggest that the model captures the correct directional structure of the motion but systematically underestimates its magnitude. This behavior is consistent with the compression effect introduced by the 256-bin discretization of the action space, and not attributable to miscalibration, as the Berkeley dataset statistics were explicitly used for denormalization.

Overall, these results indicate that, in an offline and controlled setting, the pretrained model and inference pipeline capture a meaningful portion of the demonstrated motion structure. The model is able to reproduce coherent trajectory patterns when evaluated on data closely related to its training distribution. At the same time, the need for alignment and scaling corrections reveals underlying inconsistencies in action representation and normalization. Even in an approximately in-distribution setting, the mapping between predicted actions and the physical coordinate system is not directly consistent with the recorded trajectories.This experiment validates the overall integrity of the inference pipeline and confirms that the model can produce structured and temporally coherent predictions in an offline setting. However, it also highlights that representation-level mismatches are already present prior to real-world deployment, suggesting that additional sources of error may emerge in closed-loop and real-robot conditions.

\subsection{Zero-Shot Deployment on the Real Robot}
\begin{figure}[htbp]
    \centering
    \begin{subfigure}[b]{0.45\textwidth}
        \centering
        \includegraphics[width=\textwidth]{{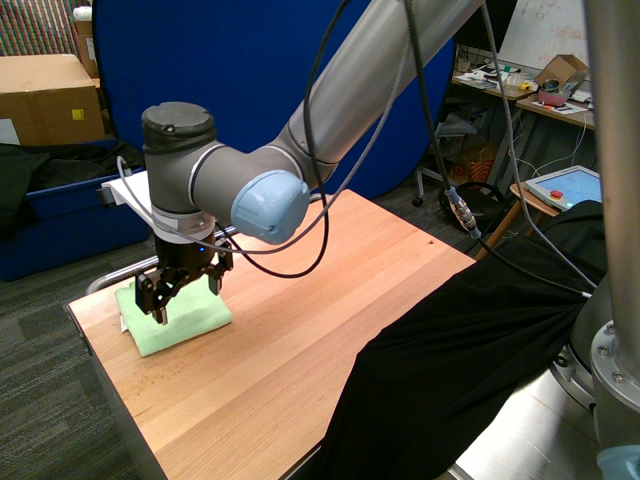}}
        \caption{Frame from Berkeley UR5 Demonstration Dataset}
        \label{fig:berk_setup}
    \end{subfigure}
    \begin{subfigure}[b]{0.45\textwidth}
        \centering
        \includegraphics[width=\textwidth, angle=-90]{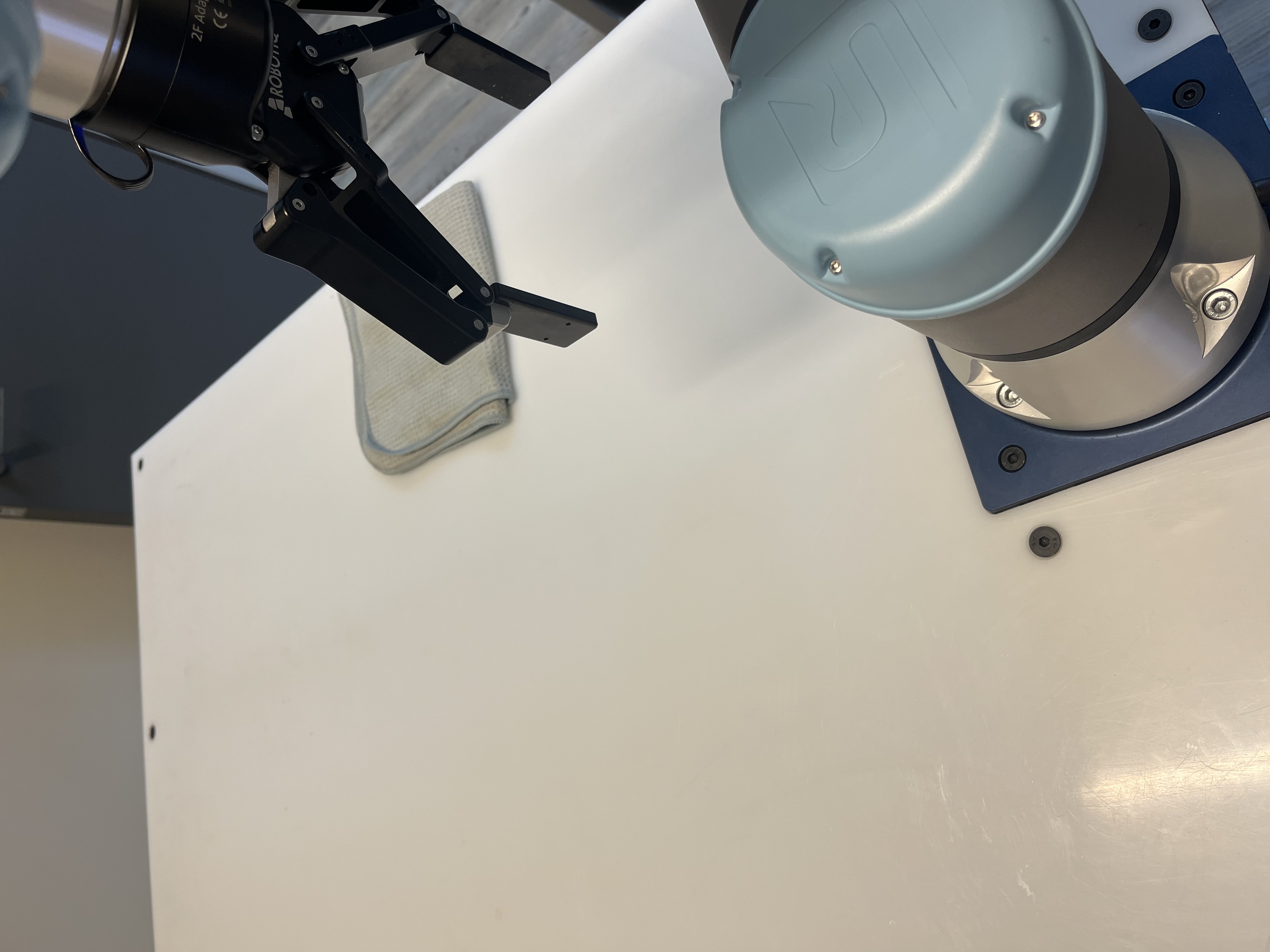}
        \caption{Reproduction of the setup in our own environment}
        \label{fig:crim}
    \end{subfigure}
    \caption{ }
    \label{fig:zeroshot}
\end{figure}

The second experiment evaluates the zero-shot behavior of OpenVLA on the UR5e platform using live camera input, with the objective of assessing its ability to generalize to a real-world setting without task-specific adaptation. Inference is performed in closed-loop on the physical system, using images acquired from the real cameras. Two experimental conditions are considered.
First, an \textit{approximately} in-distribution setup is constructed by reproducing the original dataset conditions as closely as possible. This includes a similar task structure, comparable object types, and approximate camera placement. However, exact replication is not possible due to differences in the environment (background, lighting conditions, and physical objects). Despite these discrepancies, the setup is considered approximately in-distribution, as the overall task and visual configuration remain close to the original training domain. A visual comparison between the reference dataset setup and the local setup is provided in Figure~\ref{fig:zeroshot}.

\begin{figure}[H]
    \centering    
    \includegraphics[width=8cm]{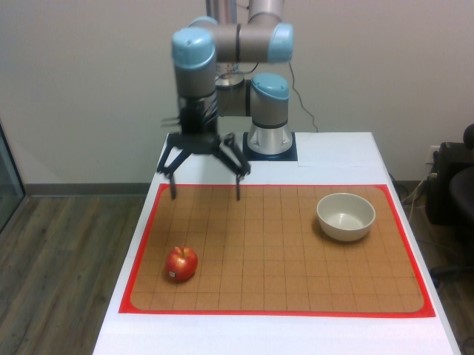}
    \caption{Frame from OOD episode from our own custom dataset.}
    \label{fig:ood}
\end{figure}

Second, an \textit{out-of-distribution} setup is introduced by designing a new scene within the local environment. This includes different object arrangements and task configurations (e.g. pick-and-place variations) that are not directly derived from the original dataset. An example of such a scene is shown in Figure~\ref{fig:ood}.

Across both conditions, the observed robot behavior is qualitatively similar. The executed motion follows a consistent direction over time, with limited variation in the predicted actions over timesteps. After a short execution period, the motion effectively stagnates. This pattern is observed in both the approximately in-distribution and out-of-distribution setups. In addition, modifications of object positions within the scene do not lead to observable changes in the executed trajectories. The resulting motions remain largely invariant despite differences in visual input. The zero-shot deployment experiment shows that OpenVLA can generate stable but largely invariant motion on the real robot, without meaningful adaptation to the environment.

\subsection{Fine-Tuning OpenVLA}

\begin{figure}[htbp]
    \centering
    \begin{subfigure}[b]{0.3\textwidth}
        \centering
        \includegraphics[width=\textwidth]{{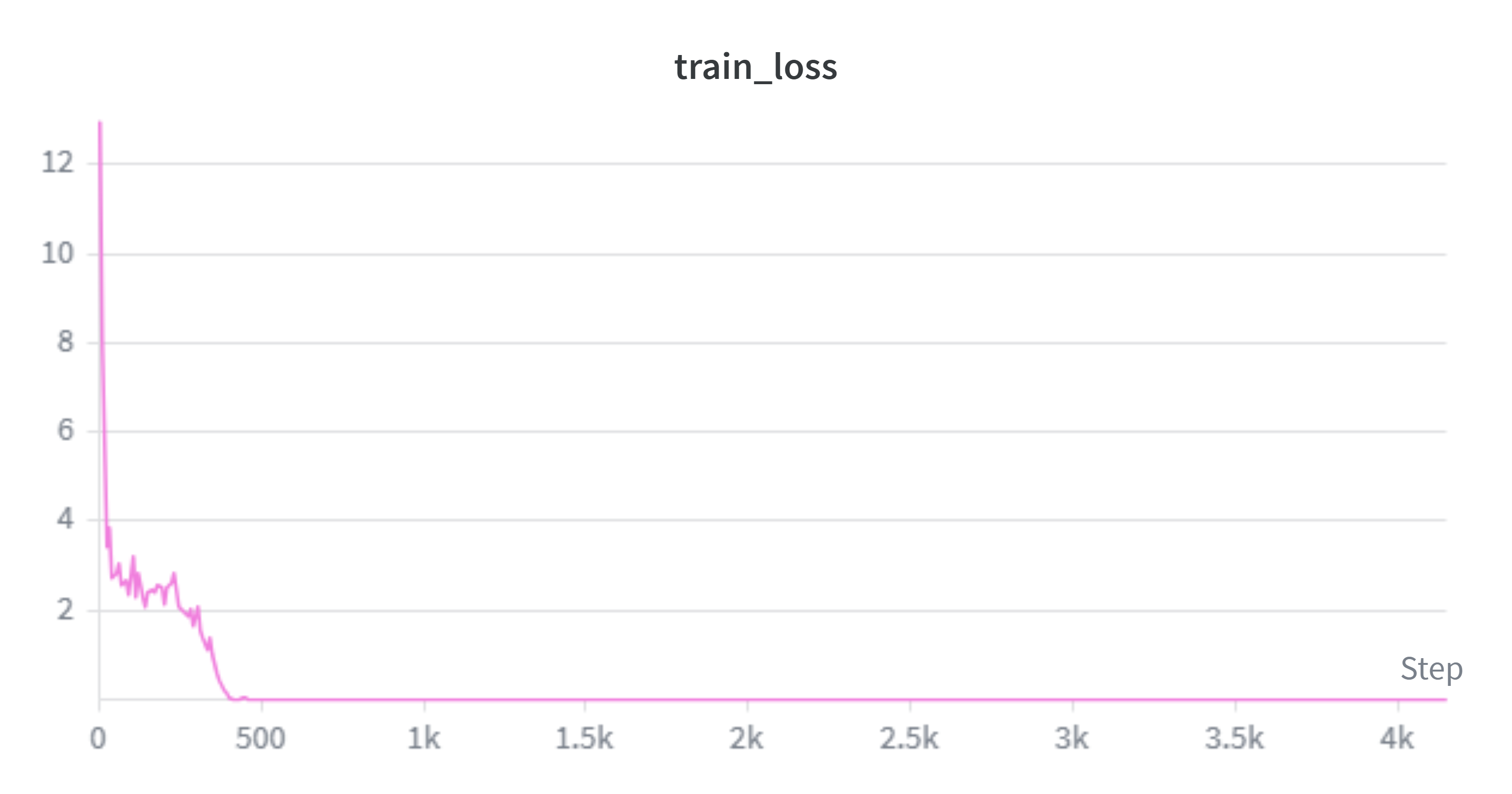}}
        \caption{Training loss}

    \end{subfigure}
    \begin{subfigure}[b]{0.3\textwidth}
        \centering
        \includegraphics[width=\textwidth]{{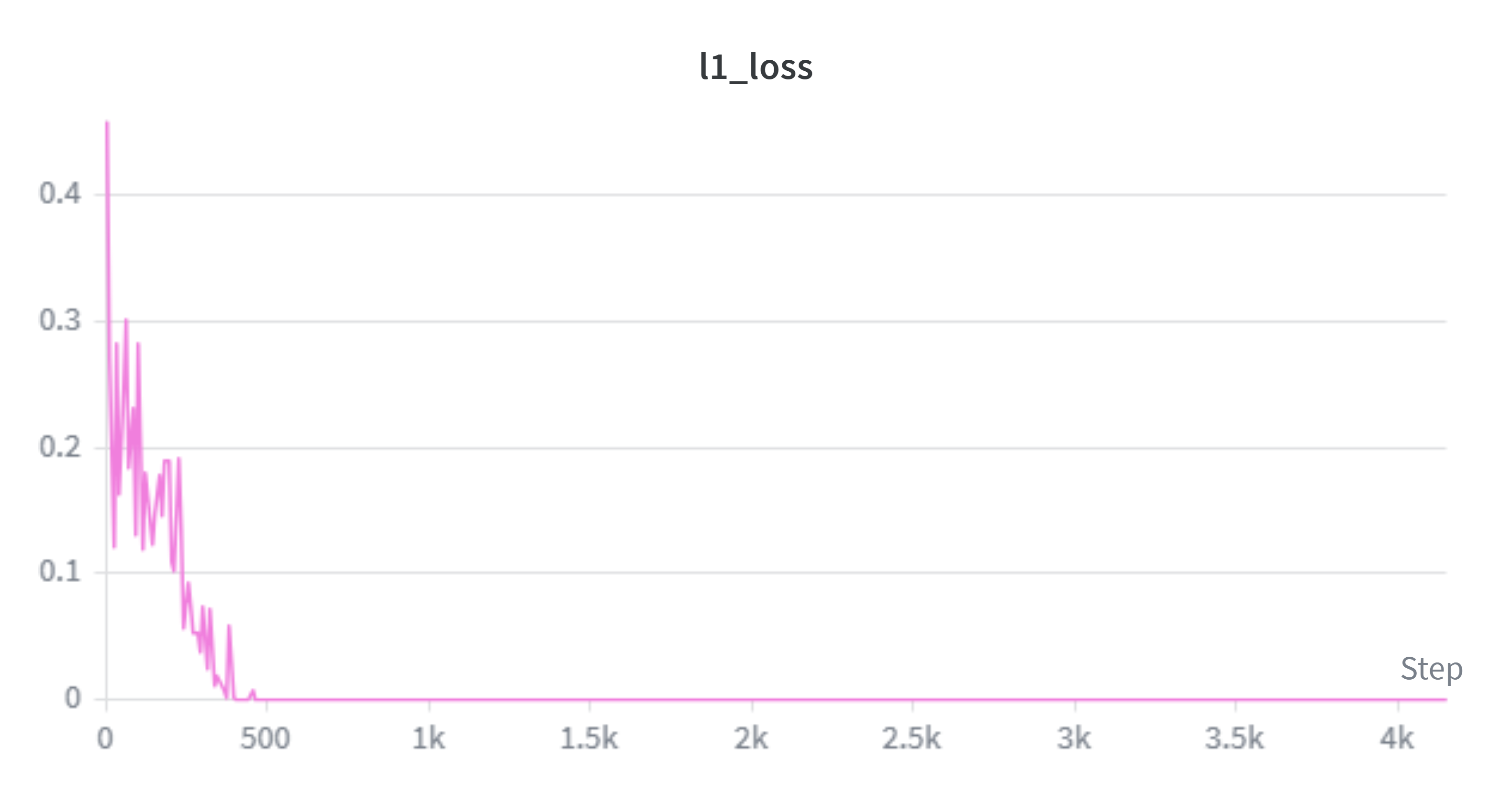}}
        \caption{L1 loss}
    \end{subfigure}
    \begin{subfigure}[b]{0.3\textwidth}
        \centering
        \includegraphics[width=\textwidth]{{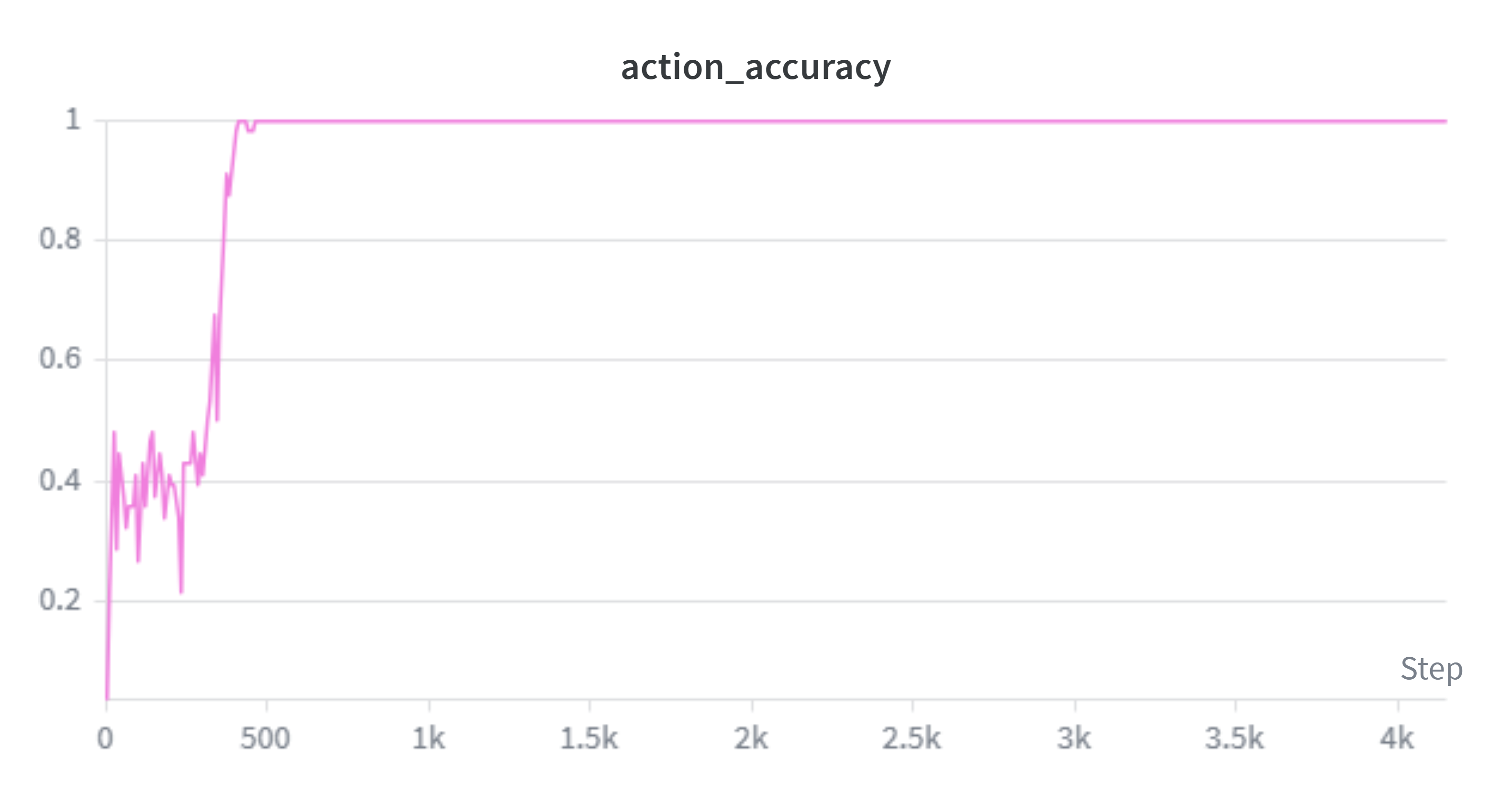}}
        \caption{Training accuracy}
    \end{subfigure}
    \caption{OpenVLA - Training curves (loss and action accuracy): This figure presents the evolution of training loss and action prediction accuracy over iterations. It documents the rapid decrease of the loss and the early saturation of the accuracy metric.}
    \label{fig:openvla-trn-metrics}
\end{figure}

The third experimental block evaluates the effect of task-specific fine-tuning on OpenVLA. Several dataset versions are used, with fine-tuning performed via LoRA and without a validation loop, following the standard OpenVLA fine-tuning pipeline. From a training perspective, all runs exhibit very rapid convergence. Training loss quickly decreases to zero, while action prediction accuracy reaches 1.0 after a small number of iterations (Figure \ref{fig:openvla-trn-metrics}). Similar training dynamics are observed across the different dataset versions. However, additional tests reveal a weak dependence of these metrics on visual input. In particular, comparable training behavior is observed when degraded or non-informative images, such as black images, are used instead of normal inputs. Furthermore, once fine-tuned, the model no longer behaves consistently on previously validated sanity-check examples: its outputs differ markedly from those of the pretrained model even when evaluated on identical inputs. Taken together, these observations indicate a strong mismatch between training indicators and effective policy behavior. Although the optimization process appears highly successful according to internal metrics, the resulting model does not preserve the expected behavior of the pretrained policy and shows limited sensitivity to the actual image content.

\begin{figure}[htbp]
    \centering
    \begin{subfigure}[b]{0.55\textwidth}
        \centering
        \includegraphics[width=\textwidth]{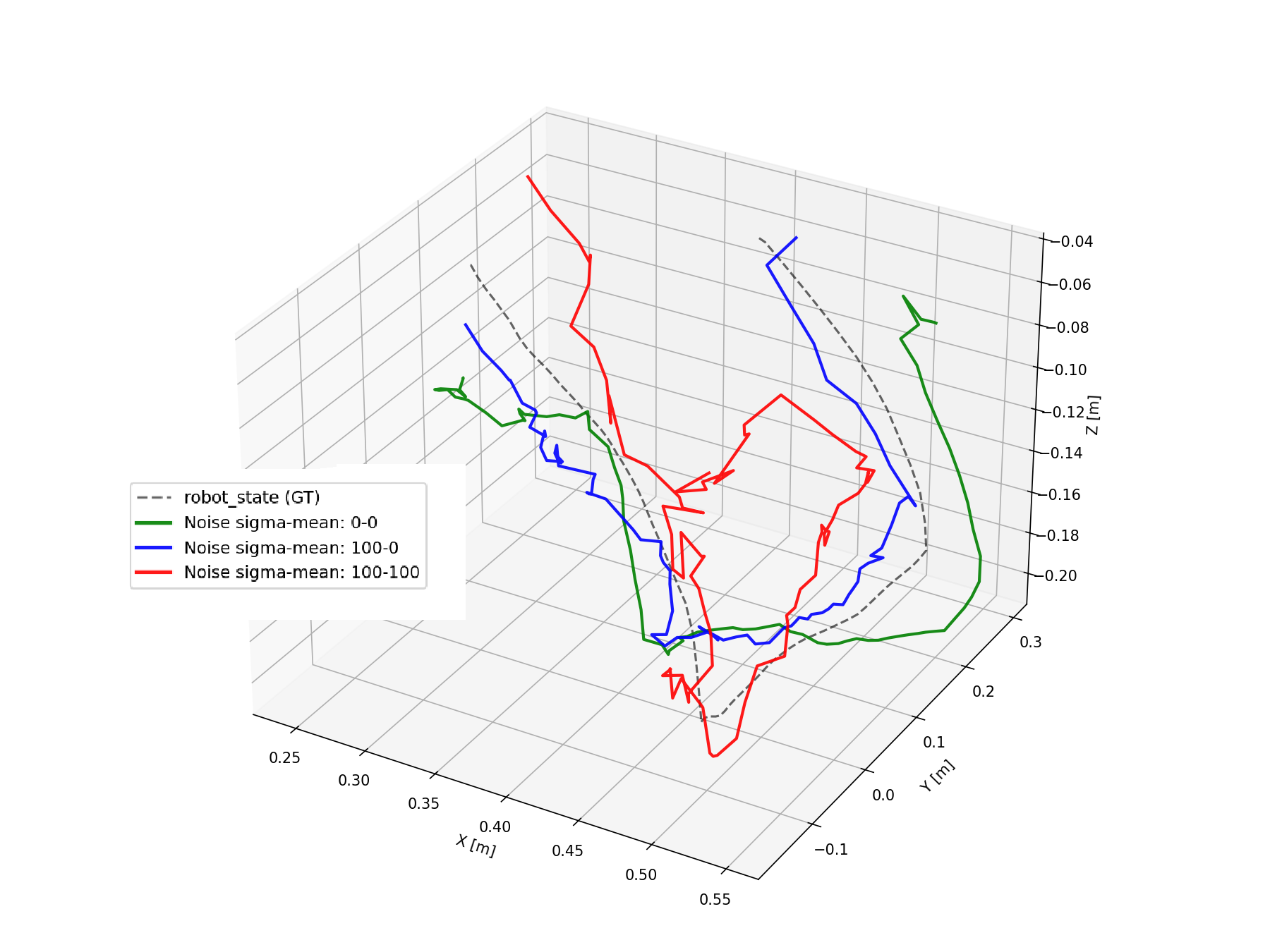}
        \caption{Robot state (GT) vs.\ OpenVLA predictions under three noise 
        conditions ($\sigma$-mean: 0-0, 100-0, 100-100).}
    \end{subfigure}

    \vspace{0.5em}

    \begin{subfigure}[b]{0.30\textwidth}
        \centering
        \includegraphics[width=\textwidth]{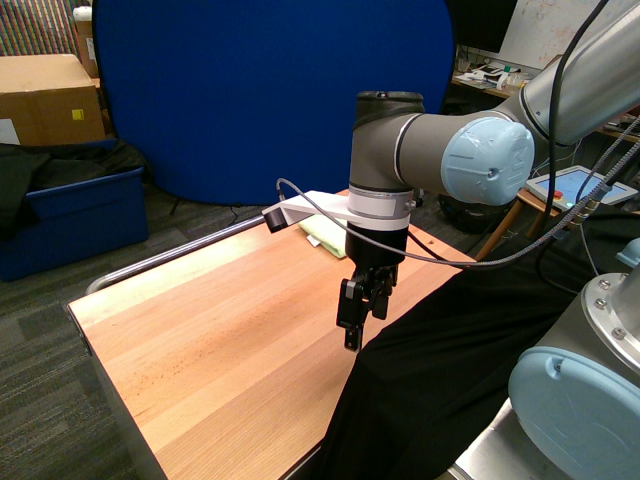}
        \caption{Static view without any degradation.}
    \end{subfigure}
    \hfill
    \begin{subfigure}[b]{0.30\textwidth}
        \centering
        \includegraphics[width=\textwidth]{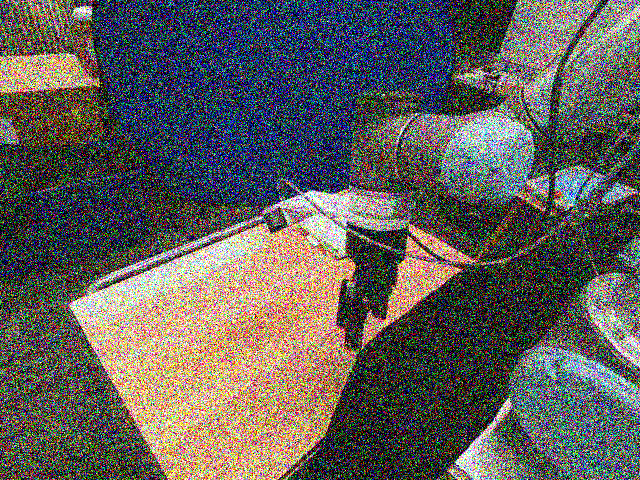}
        \caption{Static view with Gaussian noise: $\sigma=100$, $\mu=0$}
    \end{subfigure}
    \hfill
    \begin{subfigure}[b]{0.30\textwidth}
        \centering
        \includegraphics[width=\textwidth]{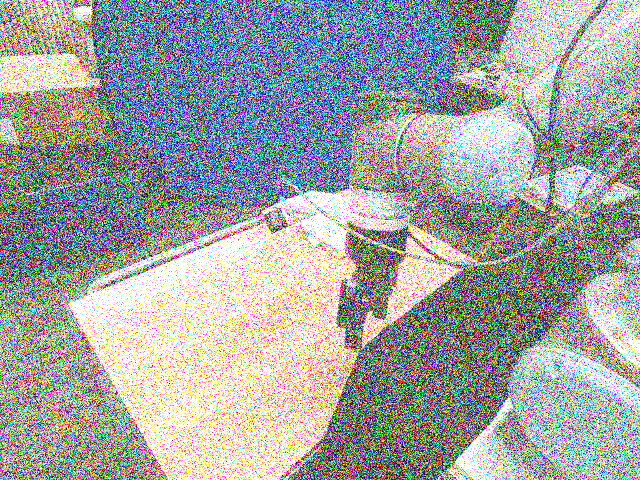}
        \caption{Static view with Gaussian noise: $\sigma=100$, $\mu=100$}
    \end{subfigure}

    \caption{Input image degradation test: predicted trajectories and corresponding 
    static frames under increasing levels of Gaussian noise injected into the 
    input images of the offline sanity-check episode 
    (\textit{sweep the green cloth}, Berkeley UR5 dataset).}
    \label{fig:image_degradation}
\end{figure}

To further characterize this insensitivity, a dedicated robustness experiment was conducted on the offline sanity-check episode ("\textit{sweep the green cloth}", Berkeley UR5 dataset). Gaussian noise was injected into the input images at three levels : no noise (sigma=0, mean=0), zero-mean noise (sigma=100, mean=0), and shifted noise (sigma=100, mean=100), and inference was run under each condition. The resulting predicted trajectories are compared in Figure \ref{fig:image_degradation} a). The most striking observation concerns the sigma=100, mean=0 condition: despite the target object (the green cloth) being very hard to distinguish in the degraded image (Figure \ref{fig:image_degradation} c)), the predicted trajectory remains structurally close to the ground truth, exhibiting only moderate deviation relative to the clean-input baseline. The condition "sigma=100, mean=100", which additionally shifts pixel intensity, produces stronger divergence, suggesting that the model is more sensitive to global luminance distribution than to local object-level content. Taken together with the black-image test reported above, these results indicate that the model's predictions are not primarily driven by task-relevant visual features, but instead appear to rely on low-level statistical properties of the image or, more likely, on structural biases inherited from the training distribution. This interpretation is consistent with independent findings reported by \cite{wang2024vlatest}, who demonstrate using a systematic fuzzing framework across seven VLA models that current VLAs exhibit insufficient robustness to visual perturbations, including lighting and object variations, and raise broader concerns about their readiness for practical deployment.

\subsection{Fine-Tuning OpenVLA-OFT}

\begin{figure}[H]
    \centering    
    \includegraphics[width=8cm]{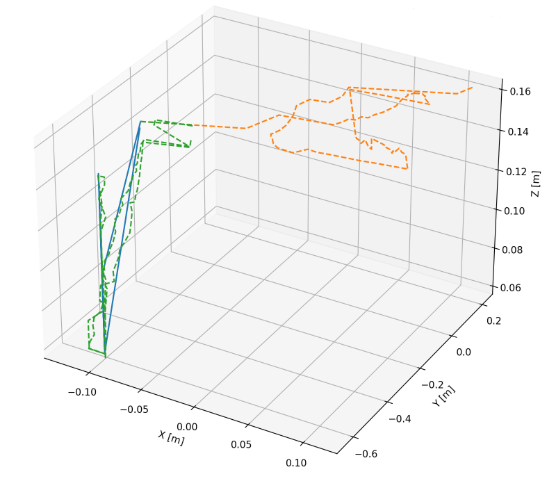}
    \caption{3D trajectory comparison (GT (blue) vs. OFT predictions (orange : closed loop and green : step-wise predictions)}
    \label{fig:oft}
\end{figure}

\begin{figure}[htbp]
    \centering
    \begin{subfigure}[b]{0.45\textwidth}
        \centering
        \includegraphics[width=\textwidth]{{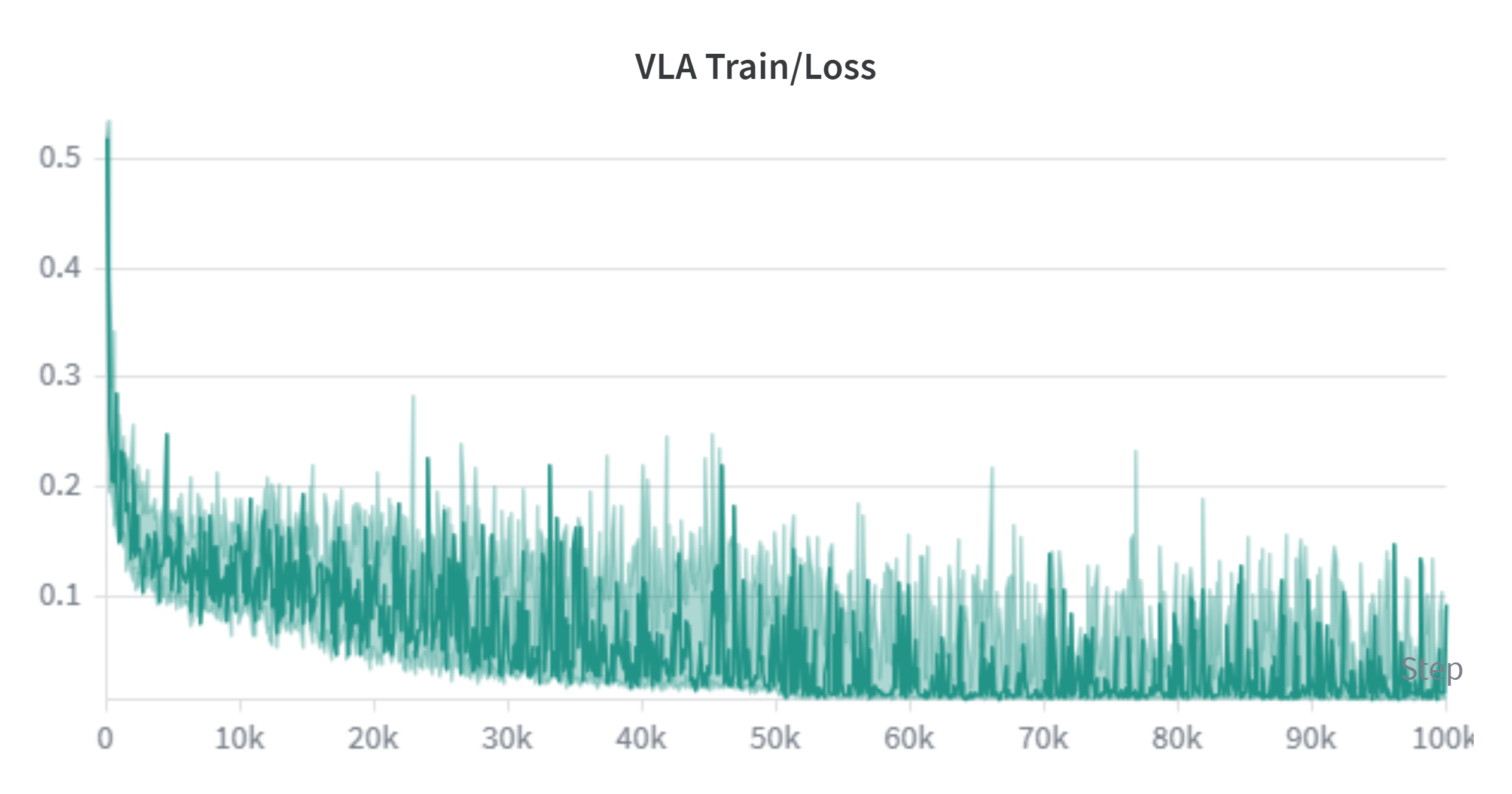}}
    \end{subfigure}
    \begin{subfigure}[b]{0.45\textwidth}
        \centering
        \includegraphics[width=\textwidth]{{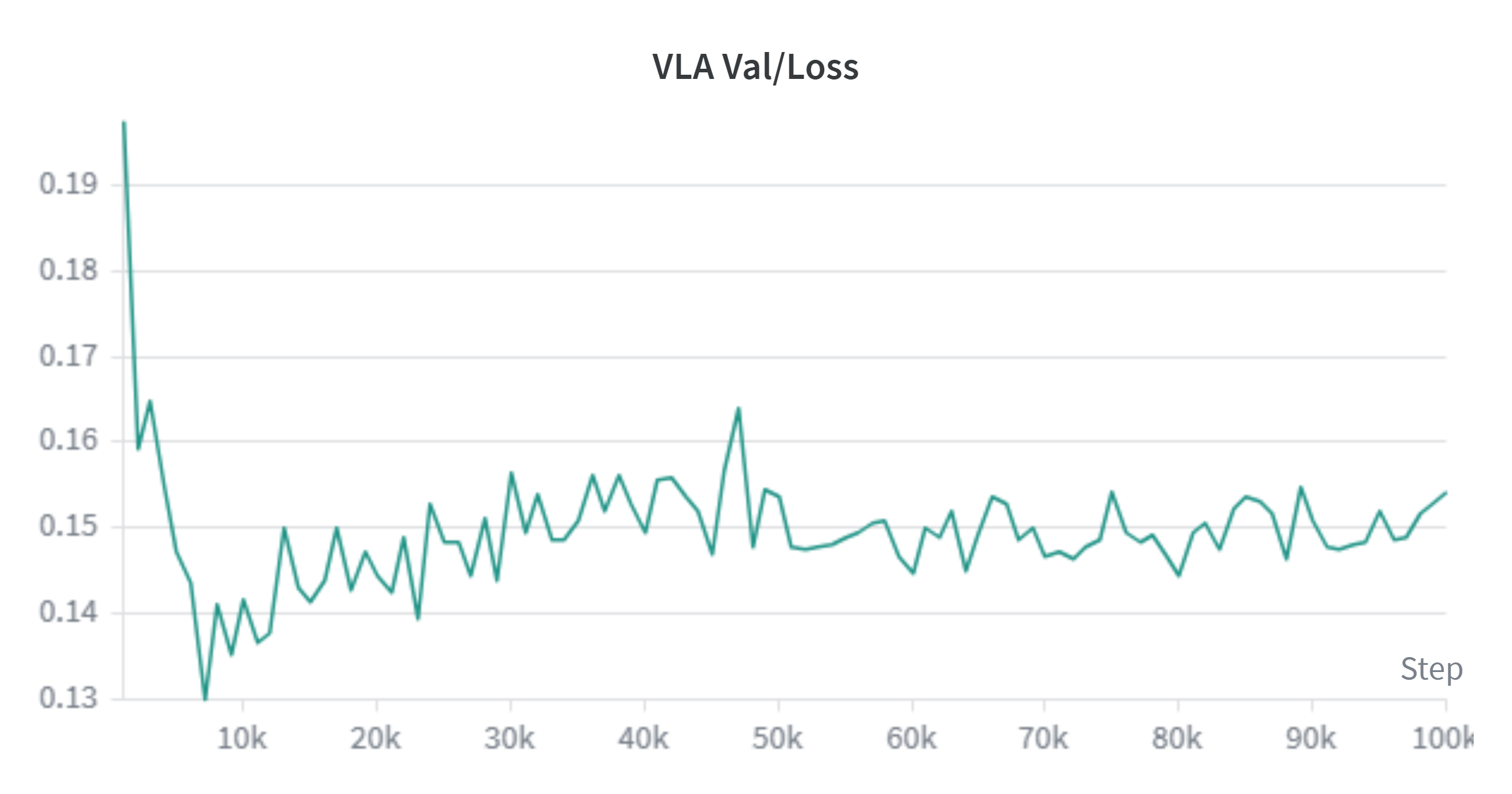}}
    \end{subfigure}
    \begin{subfigure}[b]{0.45\textwidth}
        \centering
        \includegraphics[width=\textwidth]{{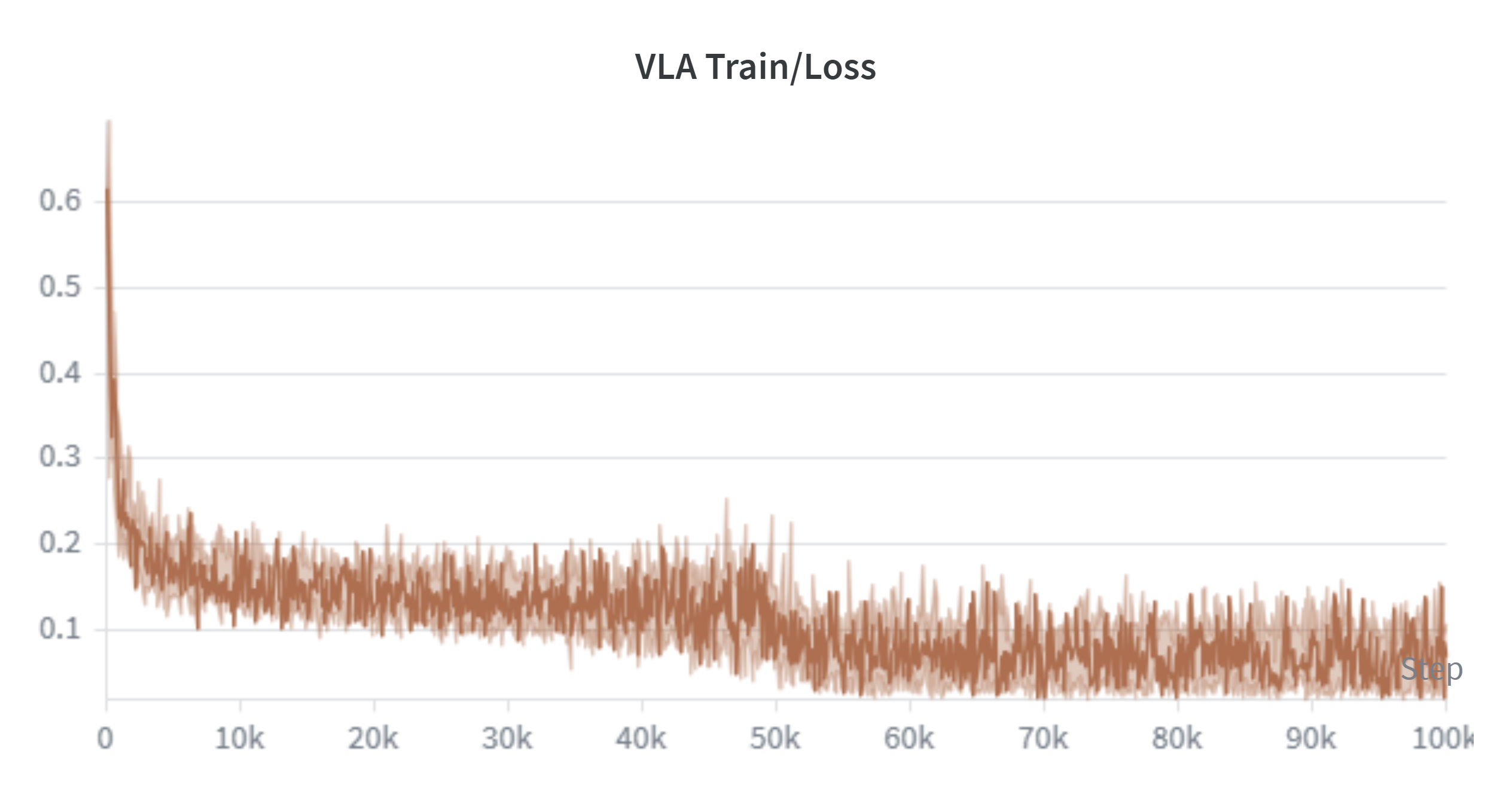}}
    \end{subfigure}
    \begin{subfigure}[b]{0.45\textwidth}
        \centering
        \includegraphics[width=\textwidth]{{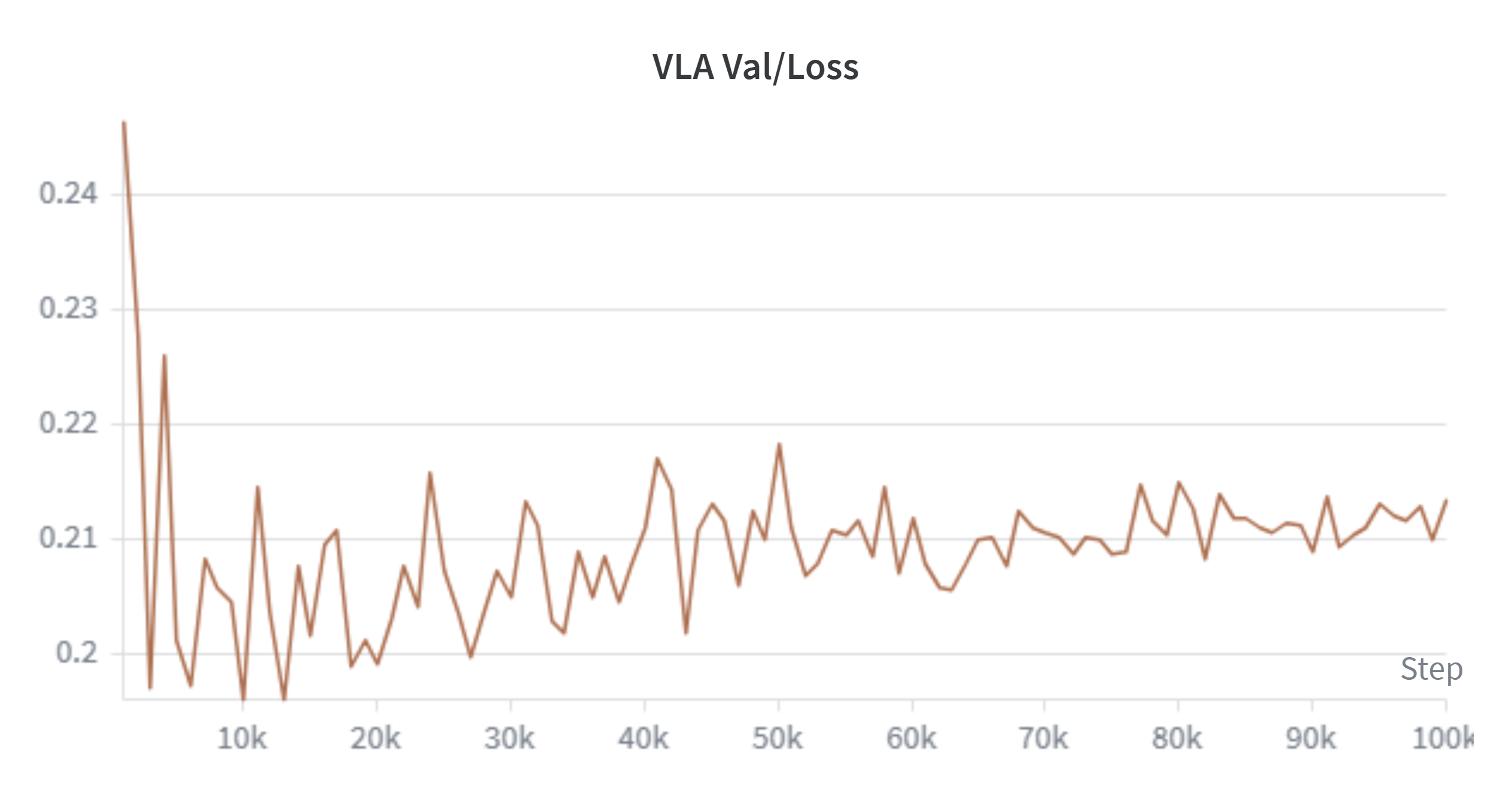}}
    \end{subfigure}
    \begin{subfigure}[b]{0.45\textwidth}
        \centering
        \includegraphics[width=\textwidth]{{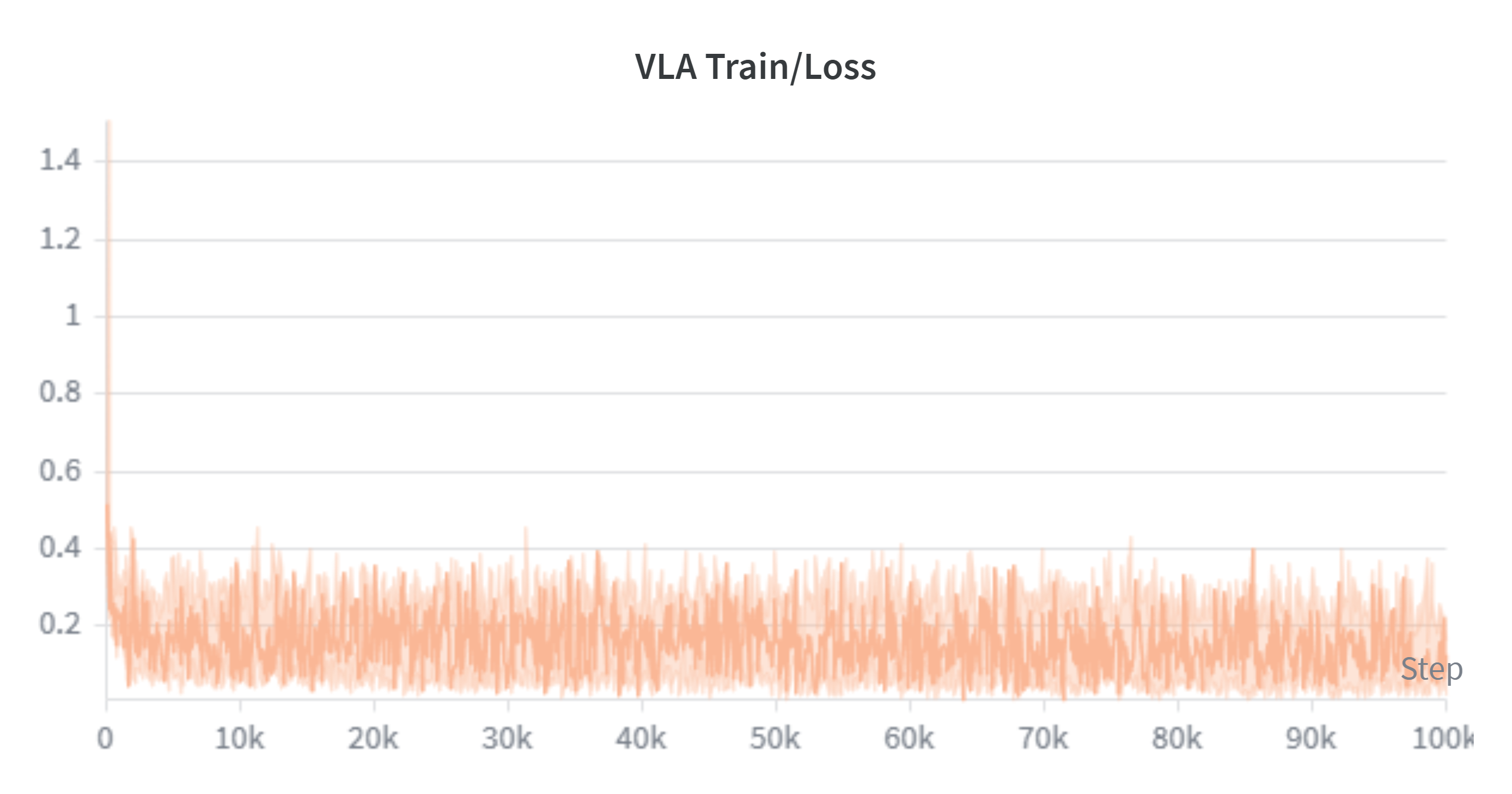}}
    \end{subfigure}
    \begin{subfigure}[b]{0.45\textwidth}
        \centering
        \includegraphics[width=\textwidth]{{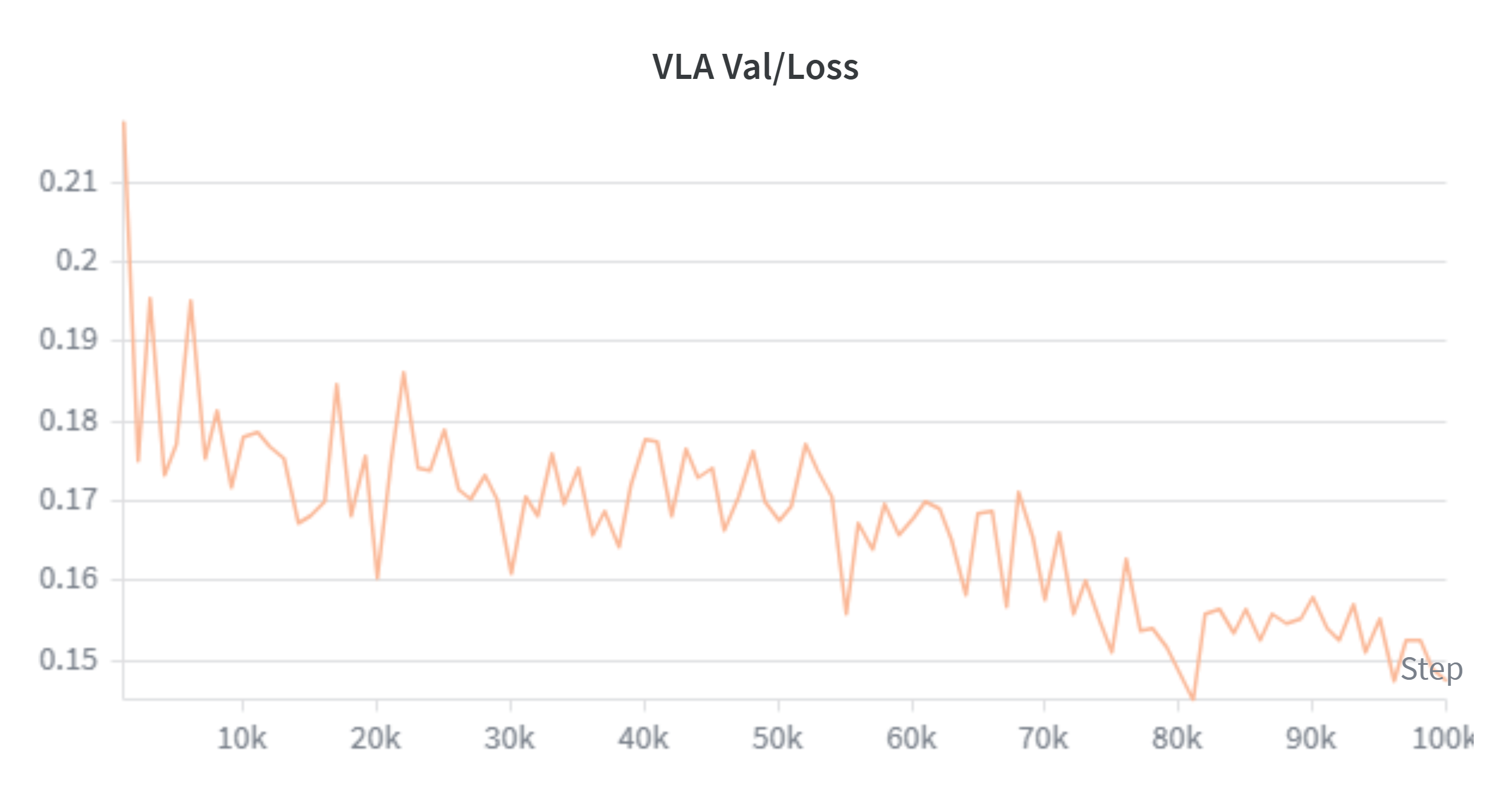}}
    \end{subfigure}
    
    \caption{OpenVLA-OFT - Loss curves (L1 loss on continuous action prediction; training and validation) on a large number of epochs. Row 1: dataset V0 (scripted). Row 2: dataset V1 (scripted,  noise added). Row 3: dataset V5-freedrive (manually guided)}
    \label{fig:openvla-oft-metrics-100kepochs}
\end{figure}

A second fine-tuning study is conducted with OpenVLA-OFT, using the OFT training formulation based on continuous action prediction, regression loss, and multi-view inputs. In contrast to OpenVLA, the training dynamics "behaves more properly" (Figure \ref{fig:openvla-oft-metrics-100kepochs}, left column). Training loss decreases steadily without abrupt collapse, and training remains stable throughout the epochs. In addition, the training script contains a validation stage, providing a more informative view of the behavior of the model during optimization (more on this later). At inference time, step-wise predictions are continuous and vary across timesteps. Over short temporal windows, these predictions generate smooth local trajectories. However, in closed-loop execution, the predicted trajectories progressively diverge from the reference trajectories, as shown in Figure~\ref{fig:oft}. These results suggest that OpenVLA-OFT produces more plausible and temporally coherent local predictions than OpenVLA, but that these improvements are not sufficient to prevent progressive deviation during closed-loop rollout. This behavior differs from the failure mode observed with OpenVLA, where training converges to degenerate solutions largely independent of visual input. In contrast, OpenVLA-OFT exhibits meaningful local predictions, but suffers from instability under recursive execution, highlighting a distinction between optimization failure and control instability.


The use of loss curves to diagnose data-related issues is a common practice in machine learning. Training runs were conducted using several datasets (V0, V1 and V5-freedrive, see Table \ref{tab:dataset_versions} for details) over a long number of epochs in order to qualitatively compare their loss curves. The datasets were selected to include a variety of characteristics (scripted, noisy, etc.). On average, each training session lasts approximately 72 hours (on four A100 GPUs). 
For dataset V0, the plateauing validation loss curve exhibits signs of overfitting. In comparison, training with the V1 episodes (roughly the same size as V0 with similar trajectories) yields the same type of curve: adding jitter to the trajectory waypoints does not have a concrete impact on training convergence.  As for the V5 dataset, it arguably contains the most natural trajectories as they have been captured while the effector was manually guided rather than being moved by a script; interestingly, the validation loss with this dataset follows a decreasing trend even after 100k epochs, indicating that the training regime is efficient and could be extended in time/epochs. Notwithstanding the fact that the analysis is qualitative and partial (e.g., no variation in hyperparameters that have a definite impact on the loss curves), it is reasonable to assume that the dataset generation procedure for V5 dataset can be reused in future work involving OpenVLA/OpenVLA-OFT (or other VLA models of interest).


\subsection{Imitation-Learning Baseline}
To support the interpretation of the previous results, a simpler supervised imitation-learning (behavior cloning) baseline is trained on the same dataset. The objective is to isolate whether the observed limitations originate from the VLA formulation itself or from the underlying data and pipeline.

\begin{figure}[H]
    \centering    
    \includegraphics[width=12cm]{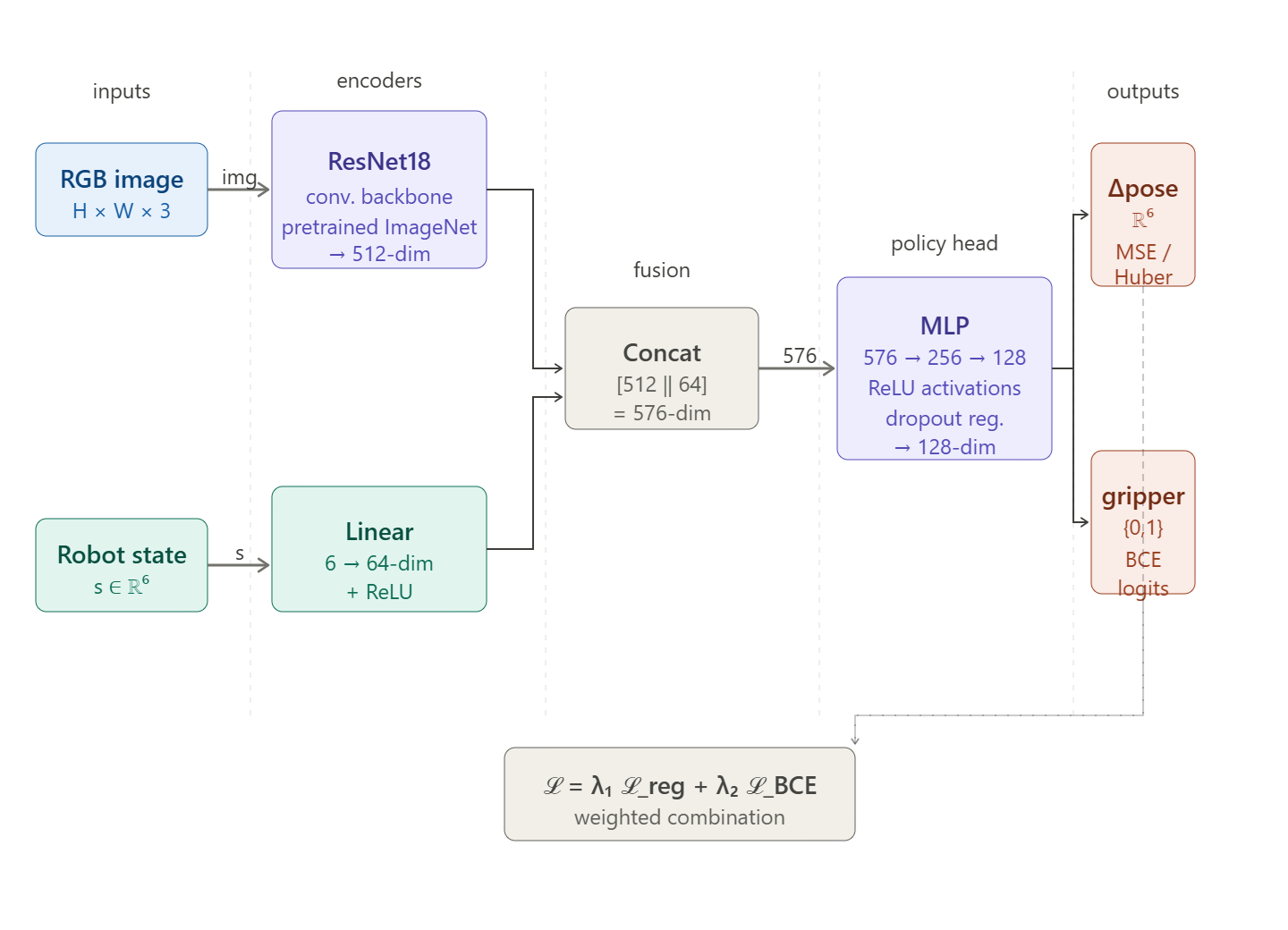}
    \caption{Architecture of the imitation-learning baseline. }
    \label{fig:imitation}
\end{figure}

The baseline policy follows a standard supervised learning formulation, where the model predicts the next action given the current observation. The input consists of an RGB image (third-person or optionally wrist) and the robot state, while the output is a 7-dimensional action vector composed of 6 continuous Cartesian deltas and a gripper command. The model architecture combines: a convolutional visual backbone (ResNet18 pretrained on ImageNet) to encode image inputs,
a state vector concatenated to the visual features, and a multi-layer perceptron (MLP) head that predicts the action vector (Figure~\ref{fig:imitation}). The training objective reflects the structure of the action space: a regression loss (MSE or Huber) is used for the 6 continuous deltas, a binary cross-entropy loss (with logits) is used for the gripper command, and the final loss is a weighted combination of both components. Then, the model is evaluated in two modes : teacher-forcing, where ground-truth states are provided at each timestep and closed-loop execution, where predicted actions are recursively fed back into the system. 

\begin{figure}[htbp]
    \centering
    \begin{subfigure}[t]{0.45\textwidth}
        \centering
        \includegraphics[width=\textwidth]{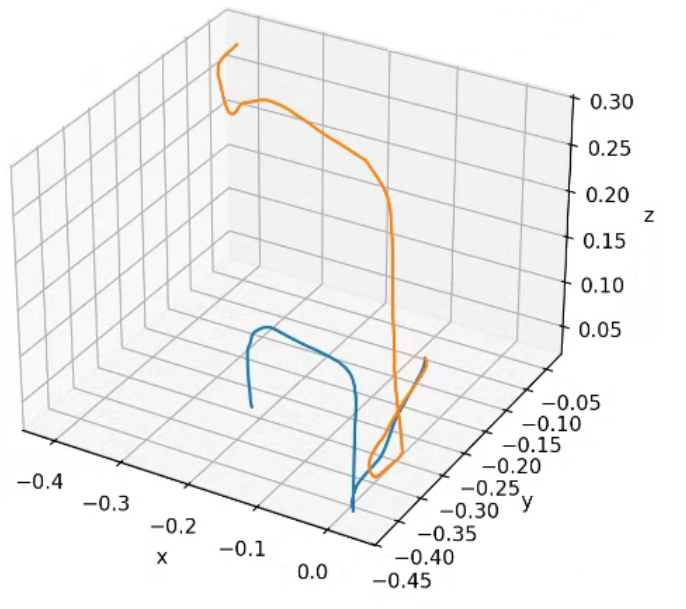}
        \caption{3D trajectory comparison : ground-truth (blue) vs imitation learning (closed-loop, orange).}
        \label{fig:baseline}
    \end{subfigure}
    \hfill
    \begin{subfigure}[t]{0.45\textwidth}
        \centering
        \includegraphics[width=\textwidth]{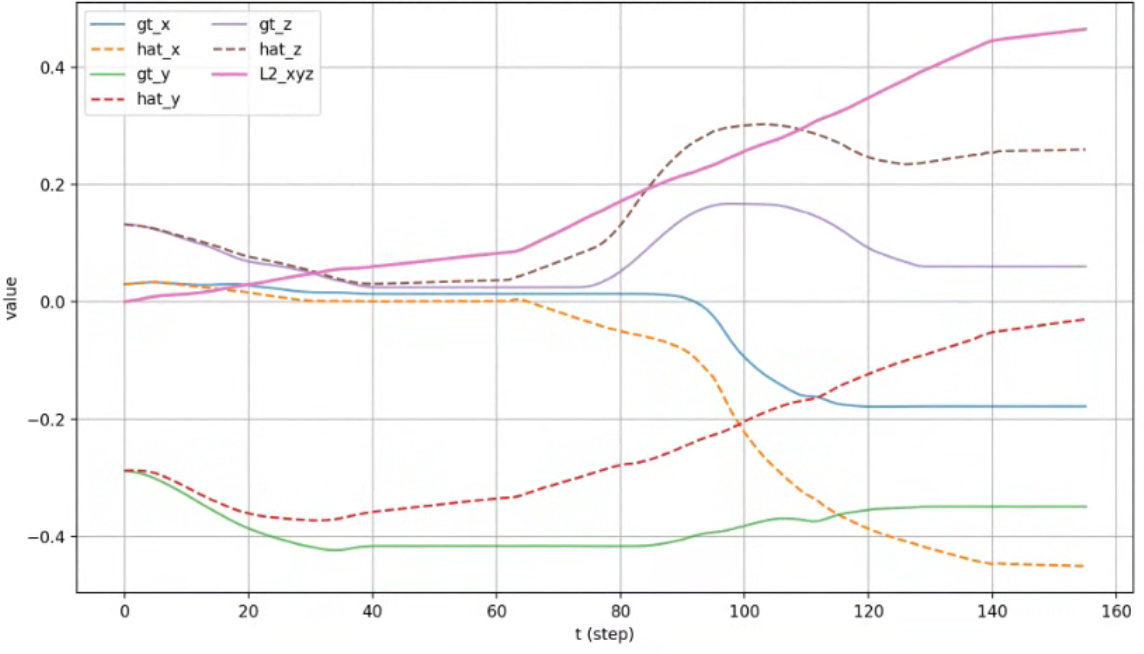}
        \caption{Time series comparison (baseline vs. GT)}
        \label{fig:time_series}
    \end{subfigure}
    \caption{(a) This figure compares the ground-truth trajectory with the imitation-learning baseline in closed-loop mode. It shows global agreement in trajectory shape with limited deviation. (b) This figure presents the Cartesian coordinates over time for both ground truth and baseline predictions. It highlights moderate deviation in closed-loop mode.}
    \label{fig:setup_in_distribution}
\end{figure}

In teacher-forcing mode, the predicted trajectories closely match the ground-truth trajectories. The model is able to reproduce both the spatial structure and the temporal evolution of the reference motion. When deployed in closed-loop, progressive deviation from the ground-truth trajectory appears over time (Figure~\ref{fig:baseline}). However, this deviation remains limited over the duration of the trajectory. Despite this drift, the global shape of the trajectory remains consistent with the reference motion (Figure~\ref{fig:time_series}). Compared to VLA-based models, the imitation-learning baseline exhibits more stable and coherent trajectory behavior on the same dataset.

\section{Discussion}
The experiments consistently reveal a fundamental gap between local prediction quality and long-horizon control stability. Across all evaluated approaches (OpenVLA, OpenVLA-OFT, and the imitation learning baseline), coherent one-step predictions do not translate into stable closed-loop trajectories. This gap is not architecture-specific: it emerges from the interaction between prediction, control, and feedback, and is consistent with known compounding error dynamics in behavior cloning \cite{dagger, closed_loop_instability}.Several approaches have been proposed to address this limitation in the context of VLA models, including on-policy reinforcement learning fine-tuning \cite{chen2025conrft}, empirically shown to improve execution robustness over SFT \cite{liu2026rlbringvlageneralization}, and RL-based post-training paradigms spanning online, offline, and test-time adaptation \cite{rl_deng_2025}. However, these approaches introduce significant additional requirements : reward design, simulation infrastructure, or large-scale real-world interaction, that fall outside the scope of the present study. The observations reported here are therefore best understood as motivating future work toward closed-loop training objectives rather than as an unaddressed gap in the literature.

Regarding the four initial hypotheses described in Section \ref{sect:hyp}, the following observations can be made: 
\begin{itemize}
    \item The bias caused by the presence of idle steps partially supports the first hypothesis. Indeed, zero-shot execution stagnates rapidly, though this may also reflect distribution mismatch rather than idle steps alone.
    \item The hypothesis about preprocessing inconsistencies that degrade performance is confirmed: the noise robustness experiment shows that the model is more sensitive to global luminance shifts than to task-relevant content, consistent with preprocessing-induced distribution artifacts.
    \item The third hypothesis regarding action representation ambiguities affecting predictions is confirmed: the offline sanity check reveals systematic amplitude compression attributable to discretization, and the transition to continuous actions in OFT improves local smoothness without resolving closed-loop instability, indicating the issue is inherent to delta-based control rather than discretization alone.
    \item Finally, the last hypothesis (dataset quality partially validatable via a simpler baseline) is confirmed: the imitation learning baseline produces more stable trajectories than either VLA model on the same data, isolating the limitation to model formulation rather than data alone.
\end{itemize}

The noise robustness experiment raises a deeper question about the information that VLA models actually use from their visual input. Predictions remain structurally coherent under severe image degradation but diverge under global luminance shifts, suggesting reliance on low-level statistical properties rather than task-relevant features. This raises unresolved questions about the role of the end-effector as a visual anchor, and about how multi-view inputs are weighted in OFT, both of which remain uncharacterized in the current literature \cite{deepvisionvla, gazeregularized}.

The comparison between OpenVLA and OFT is instructive: moving from discrete tokenized actions to continuous regression improves local coherence but does not prevent long-term divergence. The core limitation is not representational but structural : delta-based control without explicit feedback correction accumulates small systematic biases that compound over time.

More broadly, the results support the central hypothesis: real-world performance is governed by system-level alignment across the data–model–control pipeline, not by model capacity alone.

\section{Limitations and Recommendations}
The limitations identified and the recommendations formulated in this section apply specifically to OpenVLA and OpenVLA-OFT, since these are the only VLA models evaluated in this work. 

\textbf{Closed-loop evaluation must be integrated into the development cycle}. Offline metrics and teacher-forcing systematically fail to capture compounding error dynamics \cite{dagger}. Closed-loop rollouts (on the real system or in simulation) are necessary but costly; the absence of a well-aligned simulation environment limits their scalability here. While RL-based training paradigms have been proposed to explicitly address this limitation \cite{chen2025conrft, liu2026rlbringvlageneralization, rl_deng_2025}, their adoption requires reward design, simulation infrastructure, or large-scale on-robot interaction that falls outside the present scope.

\textbf{Action representation requires end-to-end consistency}. Normalization, coordinate frames, and denormalization must be strictly aligned across training and inference. However, the experiments show that even correct continuous representations do not guarantee closed-loop stability. Delta-based Cartesian control is inherently sensitive to accumulated bias. Alternative formulations (velocity control, hybrid position-feedback) warrant investigation, although they require significant pipeline modifications.

\textbf{Training must be monitored for degenerate solutions}. OpenVLA's rapid convergence combined with insensitivity to visual input indicates collapse to solutions that do not depend on perception. Validation datasets, input sensitivity tests, and diverse acquisition protocols can help detect this but do not resolve it. The training objective itself may require modifications such as the addition of multi-step consistency constraints or DAgger-style dataset aggregation \cite{dagger}.

\textbf{Hybrid approaches that combine learned policies with classical control} offer a practical path to robustness \cite{classical_robotics,feedback_control, modern_robot_control_ieee}. A VLA providing high-level proposals with a lower-level feedback controller for execution is architecturally realistic, though it introduces interface complexity and partially reduces the generality VLAs aim to provide. 

\textbf{Data quality is necessary but insufficient}. Minimizing idle steps, ensuring temporal alignment, and increasing variability are established best practices. The V5 freedrive dataset shows the most favorable training dynamics and is recommended as the baseline acquisition mode for future work. However, the absence of recovery trajectories limits the model's ability to correct deviations. Distribution coverage is as important as dataset size.

Finally, \textbf{the use of simulation environments as an intermediate validation stage} is recommended to enable systematic experimentation and debugging. A simulation aligned with the real-world setup would allow controlled testing of hypotheses related to action representation, temporal dynamics, and perception–control coupling. It would also facilitate safe evaluation of unstable policies. However, building such a simulation is non-trivial and introduces its own challenges, including the need for accurate modeling of robot dynamics, sensor characteristics, and environmental variability. Moreover, discrepancies between simulation and reality (the sim-to-real gap) may limit the transferability of conclusions drawn in simulation \cite{sim2real_survey, domain_randomization}.

\section{Conclusion}
This study evaluated the transferability of OpenVLA and OpenVLA-OFT from controlled benchmarks to a real UR5e platform. The engineering contribution is a complete, reproducible pipeline: real-robot data acquisition, RLDS-compatible dataset construction, LoRA fine-tuning workflows, and closed-loop deployment infrastructure.

The central finding is unambiguous: offline stability does not guarantee closed-loop stability. All evaluated approaches, including a simpler imitation learning baseline, exhibit trajectory drift or stagnation under recursive execution, regardless of action representation or training objective. Failures cannot be attributed to model architecture alone; they emerge from the interaction of action representation, coordinate frame consistency, temporal alignment, preprocessing, and dataset structure across the entire pipeline.

The comparison between OpenVLA and OFT confirms that improved action formulations reduce local prediction error without resolving long-horizon instability. The fundamental challenge is not prediction accuracy, but the absence of mechanisms that enforce consistency under closed-loop dynamics.

This reframes VLA deployment as a system problem. Reliable real-world performance requires co-design of data acquisition, action representation, training objectives, control strategies, and evaluation protocols. Future progress depends on moving beyond one-step prediction metrics toward frameworks that explicitly account for closed-loop behavior, error accumulation, and physical interaction.

\section*{Acknowledgments}
This work was financed by Luqia Technologies with support from the Ministry of Economy, Innovation, and Energy (MEIE) of the Government of Quebec.


\printbibliography{}

@online{octo,
  title        = {{Octo}: An Open-Source Generalist Robot Policy},
  author       = {{Octo Model Team} and Ghosh, D. and Walke, H. and
                  Pertsch, K. and Black, K. and Mees, O. and
                  Dasari, S. and Hejna, J. and Kreiman, T. and
                  Xu, C. and Luo, J. and Tan, Y. L. and
                  Chen, L. Y. and Sanketi, P. and Vuong, Q. and
                  Xiao, T. and Sadigh, D. and Finn, C. and Levine, S.},
  year         = {2024},
  eprint       = {2405.12213},
  archivePrefix = {arXiv},
  primaryClass = {cs.RO},
  doi          = {10.48550/arXiv.2405.12213},
  url          = {https://arxiv.org/abs/2405.12213}
}

@online{rt1,
      title={RT-1: Robotics Transformer for Real-World Control at Scale}, 
      author={Brohan, A. and Brown, N. and Carbajal, J. and
                  Chebotar, Y. and Dabis, J. and Finn, C. and
                  Gopalakrishnan, K. and Hausman, K. and Herzog, A. and
                  Hsu, J. and Ibarz, J. and Ichter, B. and
                  Irpan, A. and Jackson, T. and Jesmonth, S. and
                  Joshi, N. J. and Julian, R. and Kalashnikov, D. and
                  Kuang, Y. and Leal, I. and Lee, K. and
                  Levine, S. and Lu, Y. and Malla, U. and
                  Manjunath, D. and Mordatch, I. and Nachum, O. and
                  Parada, C. and Peralta, J. and Perez, E. and
                  Pertsch, K. and Quiambao, J. and Rao, K. and
                  Ryoo, M. and Salazar, G. and Sanketi, P. and
                  Sayed, K. and Singh, J. and Sontakke, S. and
                  Stone, A. and Tan, C. and Tran, H. and
                  Vanhoucke, V. and Vega, S. and Vuong, Q. and
                  Xia, F. and Xiao, T. and Xu, P. and
                  Xu, S. and Yu, T. and Zitkovich, B.},
      year={2023},
      eprint={2212.06817},
      archivePrefix={arXiv},
      primaryClass={cs.RO},
      url={https://arxiv.org/abs/2212.06817}
}

@online{rt2,
  title        = {{RT-2}: Vision-Language-Action Models Transfer Web Knowledge
                  to Robotic Control},
  author       = {Brohan, A. and Brown, N. and Carbajal, J. and
                  Chebotar, Y. and Chen, X. and Choromanski, K. and
                  Ding, T. and Driess, D. and Dubey, A. and
                  Finn, C. and Florence, P. and Fu, C. and
                  Gonzalez Arenas, M. and Gopalakrishnan, K. and
                  Han, K. and Hausman, K. and Herzog, A. and
                  Hsu, J. and Ichter, B. and Irpan, A. and
                  Joshi, N. and Julian, R. and Kalashnikov, D. and
                  Kuang, Y. and Leal, I. and Lee, L. and
                  Lee, T. E. and Levine, S. and Lu, Y. and
                  Michalewski, H. and Mordatch, I. and Pertsch, K. and
                  Rao, K. and Reymann, K. and Ryoo, M. and
                  Salazar, G. and Sanketi, P. and Sermanet, P. and
                  Singh, J. and Singh, A. and Soricut, R. and
                  Tran, H. and Vanhoucke, V. and Vuong, Q. and
                  Wahid, A. and Welker, S. and Wohlhart, P. and
                  Wu, J. and Xia, F. and Xiao, T. and
                  Xu, P. and Xu, S. and Yu, T. and Zitkovich, B.},
  year         = {2023},
  url          = {https://arxiv.org/abs/2307.15818}
}

@online{openvla,
  title        = {{OpenVLA}: An Open-Source Vision-Language-Action Model},
  author       = {Kim, M. J. and Pertsch, K. and Karamcheti, S. and
                  Xiao, T. and Balakrishna, A. and Nair, S. and
                  Rafailov, R. and Foster, E. and Lam, G. and
                  Sanketi, P. and Vuong, Q. and Kollar, T. and
                  Burchfiel, B. and Tedrake, R. and Sadigh, D. and
                  Levine, S. and Liang, P. and Finn, C.},
  journal      = {arXiv preprint arXiv:2406.09246},
  year         = {2024},
  doi          = {10.48550/arXiv.2406.09246},
  url          = {https://arxiv.org/abs/2406.09246}
}

@online{openvla_oft,
  title        = {Fine-Tuning Vision-Language-Action Models:
                  Optimizing Speed and Success},
  author       = {Kim, M. J. and Finn, C. and Liang, P.},
  journal      = {arXiv preprint arXiv:2502.19645},
  year         = {2025},
  doi          = {10.48550/arXiv.2502.19645},
  url          = {https://arxiv.org/abs/2502.19645}
}

@online{vla_survey,
  title={A Survey on Vision-Language-Action Models for Embodied {AI}},
  author={Ma, Y. and Song, Z. and Zhuang, Y. and Hao, J. and King, I.},
  journal={arXiv preprint arXiv:2405.14093},
  year={2024},
  url={http://dx.doi.org/10.1109/TNNLS.2025.3650584}
}

@article{kawaharazuka2025vla_review,
  title        = {Vision-Language-Action Models for Robotics:
                  A Review Towards Real-World Applications},
  author       = {Kawaharazuka, K. and Oh, J. and Yamada, J. and
                  Posner, I. and Zhu, Y.},
  journal      = {IEEE Access},
  year         = {2025},
  doi          = {10.1109/ACCESS.2025.3609980},
  url          = {https://arxiv.org/abs/2510.07077}
}

@online{openvla_repo,
  title={{OpenVLA} {GitHub} Repository},
  author={{OpenVLA Team}},
  howpublished={\url{https://github.com/openvla/openvla}},
  year={2024}
}

@online{openx,
  title={Open {X-E}mbodiment: Robotic Learning Datasets and {RT-X} Models},
  author={{Open X-Embodiment Collaboration} and Padalkar, Abhishek and Pooley, Acorn and Jain, Ajinkya and Bewley, Alex and Herzog, Alex and Irpan, Alex and Khazatsky, Alexander and Rai, Anant and Singh, Anikait and others},
  journal={arXiv preprint arXiv:2310.08864},
  year={2023},
  url={https://arxiv.org/abs/2310.08864}
}

@online{rlds,
  title={{RLDS}: an Ecosystem to Generate, Share and Use Datasets in Reinforcement Learning},
  author={Ramos, S. and Girgin, S. and Hussenot, L. and Vincent, D. and Yakubovich, H. and Toyama, D. and Gergely, A. and Stanczyk, P. and Marinier, R. and Harmsen, J. and Pietquin, O. and Momchev, N.},
  journal={arXiv preprint arXiv:2111.02767},
  year={2021},
  url={https://arxiv.org/pdf/2111.02767}
}

@inproceedings{dagger,
  title={A Reduction of Imitation Learning and Structured Prediction to No-Regret Online Learning},
  author={Ross, S. and Gordon, G.J. and Bagnell, J.A.},
  booktitle={Proceedings of the 14th International Conference on Artificial Intelligence and Statistics (AISTATS)},
  series={Proceedings of Machine Learning Research},
  volume={15},
  pages={627--635},
  year={2011},
  publisher={PMLR},
  note={arXiv:1011.0686},
  url={https://arxiv.org/pdf/1011.0686}
}

@book{classical_robotics,
  title={Robotics: Modelling, Planning and Control},
  author={Siciliano, Bruno and Sciavicco, Lorenzo and Villani, Luigi and Oriolo, Giuseppe},
  publisher={Springer},
  year={2009},
  isbn={978-1-84628-641-4}
}

@online{sim2real_survey,
  title={Sim-to-Real Transfer in Deep Reinforcement Learning for Robotics: a Survey},
  author={Zhao, W. and Queralta, J.P. and Westerlund, T.},
  journal={arXiv preprint arXiv:2009.13303},
  year={2020},
  url={https://arxiv.org/abs/2009.13303}
}

@inproceedings{domain_randomization,
  title={Domain Randomization for Transferring Deep Neural Networks from Simulation to the Real World},
  author={Tobin, J. and Fong, R. and Ray, A. and Schneider, J. and Zaremba, W. and Abbeel, P.},
  booktitle={2017 IEEE/RSJ International Conference on Intelligent Robots and Systems (IROS)},
  pages={23--30},
  year={2017},
  doi={10.1109/IROS.2017.8202133},
  note={arXiv:1703.06907},
  url={https://arxiv.org/abs/1703.06907}
}

@book{feedback_control,
  title={Robot Modeling and Control},
  author={Spong, Mark W. and Hutchinson, Seth and Vidyasagar, M.},
  publisher={John Wiley \& Sons},
  year={2005},
  isbn={978-0-471-64990-8}
}

@article{modern_robot_control_ieee,
  title={Robot Collisions: A Survey on Detection, Isolation, and Identification},
  author={Haddadin, Sami and De Luca, Alessandro and Albu-Schaffer, Alin},
  journal={IEEE Transactions on Robotics},
  volume={33},
  number={6},
  pages={1292--1312},
  year={2017},
  doi={10.1109/TRO.2017.2723903},
  url={https://ieeexplore.ieee.org/document/8059840}
}

@inproceedings{closed_loop_instability,
  title={Transporter Networks: Rearranging the Visual World for Robotic Manipulation},
  author={Zeng, Andy and Florence, Pete and Tompson, Jonathan and Welker, Stefan and Chien, Jonathan and Attarian, Maria and Armstrong, Travis and Florencio, Dinei and others},
  booktitle={Proceedings of the 2020 Conference on Robot Learning},
  year={2021},
  url={https://proceedings.mlr.press/v155/zeng21a.html}
}

@misc{zhou2026liberoprorobustfairevaluation,
      title={LIBERO-PRO: Towards Robust and Fair Evaluation of Vision-Language-Action Models Beyond Memorization}, 
      author={Xueyang Zhou and Yangming Xu and Guiyao Tie and Yongchao Chen and Guowen Zhang and Duanfeng Chu and Pan Zhou and Lichao Sun},
      year={2026},
      eprint={2510.03827},
      archivePrefix={arXiv},
      primaryClass={cs.CV},
      url={https://arxiv.org/abs/2510.03827}, 
}

@misc{fei2025liberoplusindepthrobustnessanalysis,
      title={LIBERO-Plus: In-depth Robustness Analysis of Vision-Language-Action Models}, 
      author={Senyu Fei and Siyin Wang and Junhao Shi and Zihao Dai and Jikun Cai and Pengfang Qian and Li Ji and Xinzhe He and Shiduo Zhang and Zhaoye Fei and Jinlan Fu and Jingjing Gong and Xipeng Qiu},
      year={2025},
      eprint={2510.13626},
      archivePrefix={arXiv},
      primaryClass={cs.RO},
      url={https://arxiv.org/abs/2510.13626}, 
}

@inproceedings{calvin_2022,
  title={CALVIN: A Benchmark for Language-Conditioned Policy Learning for Long-Horizon Robot Manipulation Tasks},
  author={Mees, Oier and Hermann, Lukas and Rosete-Beas, Erick and Burgard, Wolfram},
  booktitle={Conference on Robot Learning (CoRL)},
  year={2022},
  url={https://arxiv.org/abs/2112.03227}
}

@online{libero_2023,
  title        = {{LIBERO}: Benchmarking Knowledge Transfer for
                  Lifelong Robot Learning},
  author       = {Liu, B. and Zhu, Y. and Gao, C. and Feng, Y. and
                  Liu, Q. and Zhu, Y. and Stone, P.},
  journal      = {arXiv preprint arXiv:2306.03310},
  year         = {2023},
  doi          = {10.48550/arXiv.2306.03310},
  url          = {https://arxiv.org/abs/2306.03310}
}

@online{lora_2021,
  title         = {{LoRA}: Low-Rank Adaptation of Large Language Models},
  author        = {Edward J. Hu and
                   Yelong Shen and
                   Phillip Wallis and
                   Zeyuan Allen-Zhu and
                   Yuanzhi Li and
                   Shean Wang and
                   Lu Wang and
                   Weizhu Chen},
  year          = {2021},
  eprint        = {2106.09685},
  archivePrefix = {arXiv},
  primaryClass  = {cs.CL},
  doi           = {10.48550/arXiv.2106.09685},
  url           = {https://arxiv.org/abs/2106.09685},
  note          = {arXiv:2106.09685 [cs.CL]}
}

@article{wang2024vlatest,
  title     = {{VLATest}: Testing and Evaluating Vision-Language-Action Models for Robotic Manipulation},
  author    = {Wang, Zhijie and Zhou, Zhehua and Song, Jiayang and Huang, Yuheng and Shu, Zhan and Ma, Lei},
  journal   = {Proceedings of the ACM on Software Engineering},
  volume    = {2},
  number    = {FSE},
  articleno = {FSE073},
  year      = {2025},
  doi       = {10.1145/3729343},
  eprint    = {2409.12894},
  archivePrefix = {arXiv},
  primaryClass  = {cs.SE},
  url           = {https://arxiv.org/abs/2409.12894},
  note      = {arXiv:2409.12894v2}
}

@online{deepvisionvla,
  title         = {Look Before Acting: Enhancing Vision Foundation 
                   Representations for Vision-Language-Action Models},
  author        = {Luo, Yulin and Chen, Hao and Wu, Zhuangzhe and Sui, Bowen 
                   and Liu, Jiaming and Gu, Chenyang and Liu, Zhuoyang 
                   and Feng, Qiuxuan and Yu, Jiale and Gu, Shuo 
                   and Jia, Peng and Heng, Pheng-Ann and Zhang, Shanghang},
  year          = {2026},
  eprint        = {2603.15618},
  archivePrefix = {arXiv},
  primaryClass  = {cs.CV},
  url           = {https://arxiv.org/abs/2603.15618}
}

@online{gazeregularized,
  title         = {Gaze-Regularized Vision-Language-Action Models 
                   for Robotic Manipulation},
  author        = {Pani, Anupam and Yang, Yanchao},
  year          = {2026},
  eprint        = {2603.23202},
  archivePrefix = {arXiv},
  primaryClass  = {cs.CV},
  url           = {https://arxiv.org/abs/2603.23202}
}

@online{chen2025conrft,
  title={ConRFT: A Reinforced Fine-tuning Method for VLA Models via Consistency Policy},
  author={Chen, Yuhui and Tian, Shuai and Zhou, Yingting and Liu, Shugao and Li, Haoran, and Zhao, Dongbin},
  journal={arXiv preprint arXiv:2502.05450},
  year={2025}
}

@online{liu2026rlbringvlageneralization,
      title={What Can RL Bring to VLA Generalization? An Empirical Study}, 
      author={Jijia Liu and Feng Gao and Bingwen Wei and Xinlei Chen and Qingmin Liao and Yi Wu and Chao Yu and Yu Wang},
      year={2026},
      eprint={2505.19789},
      archivePrefix={arXiv},
      primaryClass={cs.LG},
      url={https://arxiv.org/abs/2505.19789}, 
}

@online{rl_deng_2025,
    author = {Haoyuan Deng  and Zhenyu Wu  and Haichao Liu  and Wenkai Guo  and Yuquan Xue  and Ziyu Shan  and Chuanrui Zhang  and Bofang Jia  and Yuan Ling  and Guanxing Lu  and Ziwei Wang },
    title = {A Survey on Reinforcement Learning of Vision-Language-Action Models for Robotic Manipulation},
    journal = {TechRxiv},
    volume = {2025},
    number = {1209},
    pages = {},
    year = {2025},
    doi = {10.36227/techrxiv.176531955.54563920/v1},
    URL = {https://www.techrxiv.org/doi/abs/10.36227/techrxiv.176531955.54563920/v1},
    eprint = {https://www.techrxiv.org/doi/pdf/10.36227/techrxiv.176531955.54563920/v1},
}

\end{document}